\documentclass[]{article}

\usepackage{arxiv}

\usepackage[utf8]{inputenc}
\usepackage[T1]{fontenc}  
\usepackage{hyperref}     
\usepackage{url}         
\usepackage{booktabs}     
\usepackage{amsfonts}     
\usepackage{nicefrac}     
\usepackage{microtype}      
\usepackage{lipsum}

\usepackage{floatrow}   
\usepackage{subcaption}
\usepackage{graphicx} 
\usepackage{amsmath}
\usepackage{natbib}
\usepackage{algpseudocode,algorithm,algorithmicx}

\newcommand\numberthis{\addtocounter{equation}{1}\tag{\theequation}}

\usepackage{float}

\usepackage{algpseudocode,algorithm,algorithmicx}

\newcommand{\KL}{D_\mathrm{KL}}
\newcommand{\E}{\mathbb{E}}

\author{
 Beren Millidge \\
  School of Informatics\\
  University of Edinburgh\\
  \texttt{beren@millidge.name}
   \And
    Alexander Tschantz \\
  Sackler Center for Consciousness Science\\
  School of Engineering and Informatics\\
  University of Sussex \\
  \texttt{tschantz.alec@gmail.com} \\
  \And
  Anil K Seth \\
  Sackler Center for Consciousness Science\\
   Evolutionary and Adaptive Systems Research Group\\
  School of Engineering and Informatics\\
  \And
    Christopher L Buckley \\
  Evolutionary and Adaptive Systems Research Group\\
  School of Engineering and Informatics\\
  University of Sussex\\
  \texttt{C.L.Buckley@sussex.ac.uk} 
  }
\title{Neural Kalman Filtering}

\begin{document}

\maketitle

\begin{abstract}

The Kalman filter is a fundamental filtering algorithm that fuses noisy sensory data, a  previous state estimate, and a dynamics model to produce a principled estimate of the current state. It assumes, and is optimal for, linear models and white Gaussian noise. Due to its relative simplicity and general effectiveness, the Kalman filter is widely used in engineering applications. Since many sensory problems the brain faces are, at their core, filtering problems, it is possible that the brain possesses neural circuitry that implements equivalent computations to the Kalman filter. The standard approach to Kalman filtering requires complex matrix computations that are unlikely to be directly implementable in neural circuits. In this paper, we show that a gradient-descent approximation to the Kalman filter requires only local computations with variance weighted prediction errors. Moreover, we show that it is possible under the same scheme to adaptively learn the dynamics model with a learning rule that corresponds directly to Hebbian plasticity. We demonstrate the performance of our method on a simple Kalman filtering task, and propose a neural implementation of the required equations.

\end{abstract}
\bigskip

The Bayesian Brain hypothesis has gained significant traction in cognitive neuroscience over the last two decades \citep*{knill2004bayesian,doya2007bayesian,pouget2013probabilistic}. The idea that the brain is \textit{in some sense} performing (some approximation to) Bayesian inference to solve tasks is fairly mainstream, and has received extensive experimental support \citep*{kersten2004object,kording2004bayesian,tenenbaum2006theory,angelaki2009multisensory,ernst2002humans}, although there is still significant debate on the exact nature of the computations performed, and especially about whether probabilities and probability densities such as likelihoods and priors are represented explicitly, implicitly, or not at all \citep*{sanborn2016bayesian,ma2006bayesian,ma2008spiking,pouget2000information}.

What is not as widely appreciated in the literature is that not only does the brain have to do inference on the hidden states of the world from sensory data, it must also track the changes in those states over time. This means that the brain must infer a \textit{time-varying} posterior and constantly keep it updated as the outside world evolves. This changes the problem the brain faces from that of Bayesian inference, to Bayesian filtering \citep*{kutschireiter2018nonlinear}.

The filtering problem is as follows \citep*{jaswinskistochastic,stengel1994optimal}. You have some estimated state $\hat{x}_t$, and some model of how the world evolves (the dynamics model): $\hat{x}_{t+1} = f(\hat{x}_t)$. You also receive observations $y$, and you have some model of how the observations depend on the estimated state (the observation model): $y = g(\hat{x}_t)$. Your task, then, is to compute $p(\hat{x}_{t+1} | \hat{x}_t, y_{1...t})$. In the general nonlinear case, this calculation is analytically intractable and extremely expensive to compute exactly. Some form of approximate solution is required. Two forms of approximation are generally used. The first is to approximate the model - such as by assuming linearity of the dynamics and observation models. This approach is taken by the Kalman filter. With the assumption of an initial Gaussian distribution to start with, the linear model implies that the posterior distribution will remain Gaussian for all time, and so analytic update rules for the mean and covariance can be derived .

The second method is to approximate the posterior -- usually with a set of samples (or particles). This approach is taken by the class of particle filtering algorithms which track the changing posterior by propagating the particles through the dynamics and then resampling based upon updated measurement information \citep*{arulampalam2002tutorial,gordon1993novel}. This approach can handle general nonlinear filtering cases, but suffers strongly from the curse of dimensionality. If the state-space is high-dimensional the number of particles required for a good approximation grows rapidly \citep*{doucet2000sequential}.

While there has been some work focusing on the implementation of particle filtering methods in neural circuitry \citep*{kutschireiter2015neural}, we focus on a neural implementation of the Kalman filter. There is substantial evidence that the brain is capable of Bayes-optimal integration of noisy measurements, and is apparently in possession of robust forward models both in perception \citep*{zago2008internal,simoncelli2009optimal} and motor control\citep*{munuera2009optimal, gold2003influence,todorov2004optimality}. \citet*{de2013kalman} have even shown that a Kalman filter successfully fits psychomotor data on visually guided saccades and smooth pursuit movement, although they remain agnostic on how it may be implemented in the brain.

In this paper, we show that it is possible to derive a gradient descent approximation to the analytical Kalman filter equations, which results in relatively simple update equations that are biologically plausible and can be embedded in a straightforward neural architecture of rate-coded integrate-and-fire neurons. This approach eschews any complicated linear algebra or explicit computation of the Kalman gain matrix. Moreover, it also provides for adaptively learning the coefficients of the dynamics and measurements models through Hebbian plasticity. 
The paper comprises three sections. First, we provide a general introduction to Kalman filtering, and derive our gradient descent algorithm.Secondly, we propose how the resulting equations can be translated into a neural architecture of linear integrate-and-fire neurons with Hebbian plasticity. Finally, we compare the performance of our filter to exact analytical Kalman filtering on a simple simulated tracking task.

\subsection{Related Work}

Several earlier works have also tried to approach the problem of Kalman filtering in the brain. \citet*{wilson2009neural} repurpose a line attractor network and show that it recapitulates the dynamics of a Kalman filter in the regime of low prediction error. However their model only works for a single dimensional stimulus, does not encode uncertainty, and also only works when a linearisation around zero prediction error holds. \citet*{deneve2007optimal} encode Kalman filter dynamics in a recurrent attractor network. Their approach however encodes stimuli by means of basis functions, which leads to an exponentially growing number of basis functions required to tile the space as the dimensionality of the input grows. In our approach, neurons directly encode the mean of the estimated posterior distribution, which means that our network size scales linearly with the number of dimensions. Our gradient method also completely eschews the direct computation of the Kalman gain, which simplifies the required computations significantly. Additionally, \citet*{beck2007probabilistic} show that probabilistic population coding approaches can compute posteriors for the exponential family of distributions of which the Gaussian distribution is a member. However, no explicitly worked out application of the population coding approach to Kalman filtering exists, to our knowledge.

Our model is closely related to predictive coding approaches to brain function \citep*{friston2005theory,rao1999predictive,spratling2017review,spratling2008reconciling}. The predictive coding and Kalman Filtering update rules have been compared explicitly in the context of control theory \citep{baltieri2020kalman}, although without any precise statement of their mathematical relationship. In some works \citep{friston2008variational, friston2008hierarchical}, it has been claimed that predictive coding and Kalman filtering are equivalent. Interpreted strictly, this claim is false. In this paper, in effect, we make explicit and precise the relationship between predictive coding and Kalman filtering. Specifically, predictive coding and Kalman filtering optimize the same Bayesian objective (with slightly different interpretations) which is convex in the linear, Gaussian case. The Kalman Filter then derives the analytical solution to this convex minimization problem, while the predictive coding updates arise from a gradient descent on this objective, thus leading to important differences in the actual update rule. In Appendix B, we demonstrate the relationship with predictive coding mathematically.


Finally, our model can be extended straightforwardly to arbitrary nonlinear dynamics simply by changing the dynamics and observation models to include arbitrary nonlinear functions -- i.e. $x_{t+1} = f(x_t, u_t)$ and $y_t = g(x_t)$. The only effect this has on the functioning of our algorithms is introducing nonlinear derivative terms such as $\partial \frac{f(x_t, u_t)}{\partial x_t}$ into the update equations. This renders our approach a gradient descent on the Extended Kalman Filter (EKF) \citep{ribeiro2004kalman} objective, which the actual EKF algorithm solves analytically by simply using a local linearisation of the dynamics around the current state. In some cases, such as the simple activation functions used in artificial neural networks, these nonlinear derivative terms are often relatively straightforward and may be computed in a local and biologically plausible manner. Further nonlinear refinements of the Kalman Filter have been proposed, such as the Unscented Kalman Filter \citep{wan2000unscented}, and whether our gradient based approach can be extended to such algorithms remains an interesting avenue for future research.

\section{Kalman Filtering}

The Kalman Filter solves the general filtering problem presented above by making two simplifying assumptions. The first is that both the dynamics model and the observation model are linear. The second assumption is that noise entering the system is white and Gaussian. This makes both the prior and likelihoods Gaussian. Since the Gaussian distribution is a conjugate prior to itself, this induces a Gaussian posterior, which can then serve as the prior in the next timestep. Since both prior and posterior are Gaussian, filtering can continue recursively for any number of time-steps without the posterior growing in complexity and becoming intractable. The Kalman filter is the Bayes-optimal solution provided that the assumptions of linear models and white Gaussian noise are met \citep*{kalman1960new}. The Kalman Filter, due to its simplicity and utility is widely used in engineering, time-series analysis, aeronautics, and economics \citep*{grewal2010applications,leondes1970theory,schneider1988analytical,harvey1990forecasting}.

The Kalman Filter is defined upon the following linear state-space \footnote{For simplicity, the model is presented in discrete time. The continuous time analogue of the Kalman filter is the Kalman-Bucy filter \citep*{kalman1961new}. Generalization of this scheme to continuous time is an avenue for future work.}
\begin{flalign*}
    x_{t+1} &= Ax_t + Bu_t + \omega & \\
    y_{t+1} &= Cx_{t+1} + z_t \numberthis
\end{flalign*}
Where $x_t$ represents the hidden or internal state at time t. $u_t$ is the control - or known inputs to the system - at time t. Matrices A,B, and C parametrize the linear dynamics or observation models, and $\omega$ and $z$ are both zero-mean white noise Gaussian processes with covariance $\Sigma_\omega$ and $\Sigma_z$, respectively. Since the posterior $p(x_{t+1}|y_{1...t}, x_{t})$ is Gaussian, it can be represented by its two sufficient statistics -- the mean $\mu$ and covariance matrix $\Sigma_x$.

Kalman filtering proceeds by first "projecting" forward the current estimates according to the dynamics model. Then these estimates are "corrected" by new sensory data. The Kalman filtering equations are as follows:
\newline
\textbf{Projection:}
\begin{flalign*}
    & \hat{\mu}_{t+1} = A\mu_t + Bu_t  &\\
    & \hat{\Sigma}_x(t+1) = A\Sigma_x(t) A^T + \Sigma_\omega \numberthis
\end{flalign*}
\textbf{Correction}
\begin{flalign*}
    & \mu_{t+1} = \hat{\mu}_{t+1} + K(y_{t+1} - C\hat{\mu}_{t+1}) & \\
    & \Sigma_x(t+1) = (I - K)\hat{\Sigma}_x(t+1) \\
    & K = \hat{\Sigma}_x(t+1)C^T[C\hat{\Sigma}_x(t+1)C^T + \Sigma_z]^{-1} \numberthis
\end{flalign*}

Where $\mu_t$ and $\Sigma_x(t)$ are the mean and variance of the estimate of the state $x$ at time t, and K is the Kalman gain matrix. Although these update rules provide an analytically exact solution to the filtering problem, the complicated linear algebra expressions, especially that for the Kalman gain matrix K, make it hard to see how such equations could be implemented directly in the brain. 

We therefore take a different approach, which utilizes the fact that the filtering equations can be derived directly from Bayes' rule. The mean of the posterior distribution is also the MAP (maximum-a-posteriori) point, since a Gaussian distribution is unimodal. Thus, to estimate the new mean, we simply have to estimate,
\begin{flalign*}
     \underset{\hat{x}_{t+1}}{arg max} \, \,  p(\hat{x}_{t+1} | y_{t+1}, \hat{x}_t) &\propto \underset{\hat{x}_{t+1}}{argmax} \, p(y_{t+1} |\hat{x}_{t+1})p(\hat{x}_t+1 | \hat{x}_t)  &\\
    &= \underset{\hat{x}_{t+1}}{arg max} \, N(y_{t+1};C\hat{x}_{t+1}, \Sigma_z)N(\hat{x}_{t+1}; A\hat{x}_t + Bu_t, \Sigma_\omega) \\
    &= \underset{\mu_{t+1}}{arg max} \, \frac{1}{Z}exp(-(y - C\mu_{t+1})^T\Sigma_Z(y - C\mu_{t+1}) \\ &+ (\mu_{t+1} - A\mu_t - Bu_t)^T\hat{\Sigma}_x(\mu_{t+1} - A\mu_t - Bu_t) \\
    &= \underset{\mu_{t+1}}{arg min} \, -(y - C\mu_{t+1})^T\Sigma_Z(y - C\mu_{t+1}) + (\mu_{t+1} - A\mu_t - Bu_t)^T\hat{\Sigma}_x(\mu_{t+1} - A\mu_t - Bu_t) \numberthis
\end{flalign*}

The first step is just a writing out of Bayes' rule. Next the fact that both likelihood and prior are Gaussian is used to write these as Gaussian densities. In the second line, the algebraic form of the Gaussian density is substituted and we have switched the maximization variable to $\mu_{t+1}$ due to the fact that the maximum of a Gaussian is its mean. In the next line, we minimize the log probability instead of maximizing the probability, which gets rid of the exponential and the normalizing constant (which can be computed analytically since the posterior is Gaussian). \footnote{The log transformation is valid under maximization/minimization since the log function is monotonic}.

The result is a convex optimization problem which can be solved analytically to give the Kalman filter equations shown above (see Appendix A for a full derivation). However, the optimization problem can also be solved by gradient descent. Moreover, the gradients required give rise to simple biologically plausible learning rules for networks of integrate-and-fire neurons, and even allow for the adaptive computation of the coefficients of the A, B, and C matrices using nothing more than third-factor Hebbian plasticity. This means that a close approximation of Kalman filtering can be achieved in the brain without complex linear algebra.

Derivation of the gradients proceeds as follows. Our objective is the minimization of the loss function $\mathcal{L}$,
\begin{flalign*}
    &\underset{\hat{x}_{t+1}}{arg max} \, p(y_{t+1} |\hat{x}_{t+1})p(\hat{x}_{t+1} | \hat{x}_t) = argmin_{\mu_{t+1}} \mathcal{L} & \\
    \mathcal{L} &= -(y - C\mu_{t+1})^T\Sigma_Z(y - C\mu_{t+1}) + (\mu_{t+1} - A\mu_t - Bu_t)^T\hat{\Sigma}_x(\mu_{t+1} - A\mu_t - Bu_t) \\
    &= -y^T\Sigma_z y + 2\mu_{t+1}^TC^T\Sigma_z C\mu_{t+1} - \mu_{t+1}^TC^T\Sigma_z y +\mu_{t+1}^T\Sigma_x\mu_{t+1} -2\mu_{t+1}\Sigma_x A\mu_t -2\mu_{t+1}\Sigma_x Bu_t \numberthis
\end{flalign*}

We proceed by taking derivatives with respect to $\mu_{t+1}$.
\begin{flalign*}
    \frac{dL}{d\mu_{t+1}} &= 2C^T\Sigma_z y - (C^T \Sigma_z C + C^T\Sigma_z^T C)\mu_{t+1} + (\Sigma_x + \Sigma_x^T)\mu_{t+1} - 2\Sigma_x A\mu_t - 2\Sigma_x Bu_t & \\
    &= 2C^T\Sigma_z C\mu_{t+1} - 2C^T\Sigma_z C\mu_{t+1} + 2\Sigma_x \mu_{t+1} - 2\Sigma_x A\mu_t - 2\Sigma_x Bu_t \\
    &= -C^T \Sigma_z[y - C\mu_{t+1}] + \Sigma_x[\mu_{t+1} - A\mu_t - B\mu_t] \\
    &= -C^T \Sigma_z \epsilon_z +  \Sigma_x \epsilon_x \numberthis
\end{flalign*}
Where $\epsilon_z = y - C\mu_{t+1}$ and $\epsilon_x = \mu_{t+1} - A\mu_t - Bu_t$.
This means that the Kalman filter gradient is, in effect, an update scheme relying on variance weighted prediction errors which can be computed with only linear operations of weighted subtraction and addition.

It is also possible to adaptively learn the coefficients of the A,B, and C matrices by gradient descent by taking gradients of the loss function with respect to these variables. We go through each in turn.
\begin{flalign*}
    \frac{dL}{dA} &= \frac{d}{dA}[-2\mu_{t+1}^T \Sigma_x A\mu_t + \mu_t^T A^T \Sigma_x A\mu_t + \mu_t^T A^T \Sigma_x Bu_t + u_t^TB^T \Sigma_x A \mu_t] & \\
    &= -2\Sigma_x^T \mu_{t+1} \mu_t^T + \Sigma_x^T A\mu_t \mu_t^T \Sigma_x A \mu_t\mu_t^T + \Sigma_x Bu_t\mu_t^T + \Sigma_x^T Bu_t\mu_t^T \\
    &= -\Sigma_x[\mu_{t+1} - A\mu_t - Bu_t]\mu_t^T \\
    &= -\Sigma_x \epsilon_x \mu_t^T \numberthis
\end{flalign*}

This is a simple Hebbian update rule between the prediction errors and the currently expected estimate. 
A similar derivation is possible with respect to the B matrix.
\begin{flalign*}
    \frac{dL}{dB} &= \frac{dL}{dB}[2u_t^TB^T\Sigma_x A\mu_t + u_t^TB^T\Sigma_x Bu_t - 2\mu_{t+1}^T \Sigma_x Bu_t] & \\
    &= (\Sigma_x + \Sigma_x^T)Bu_t u_t^T + 2\Sigma_x A \mu_t u_t^T - 2 \Sigma_x \mu_{t+1} u_t^T \\
    &= - \Sigma_x[\mu_{t+1} - A\mu_t - Bu_t]u_t^T \\
    &= - \Sigma_x \epsilon_x u_t^T \numberthis
\end{flalign*}
Which is another simple Hebbian update rule, between the dynamical prediction errors and the control variables.

It is also possible to take derivatives with respect to the C observation matrix.
\begin{flalign*}
    \frac{dL}{dC} &= \frac{dL}{dC}[-2\mu_{t+1}^TC^TRy + \mu_{t+1}^T C^T R C \mu_{t+1}] &\\
    &= -2Ry\mu_{t+1}^T + 2RC\mu_{t+1}\mu_{t+1}^T \\
    &= -R[y - c\mu_{t+1}]\mu_{t+1}^T \\
    &= -R\epsilon_y \mu_{t+1}^T \numberthis
\end{flalign*}

Which is again a simple Hebbian update rule between the sensory prediction errors and the estimated state.  In the next section we show how these equations can be implemented in a biologically plausible neural architecture.

\section{Neural Implementation}

Our neural implementation assumes three distinct subpopulations of neurons. The first set of neurons represent the best estimate $\mu_{t+1}$ at any point in time. The other two populations represent either the dynamical prediction errors $\epsilon_x$ or the sensory prediction errors $\epsilon_y$. 
It is important to note that this scheme is only one possible equivalent representation, and is a fairly direct translation of the mathematics into neural implementation. It is possible to implement the same scheme in a variety of different neural architectures.

Sensory data $y_t$ enters the model at the bottom and is projected upwards to the sensory prediction error neurons via driving feed-forward afferent connections. The connectivity pattern of these connections is sparse (in our model one-to-one). The sensory prediction errors become variance weighted through lateral connections which encode the matrix R. These prediction errors are transmitted upwards through diffuse inhibitory connections embodying the matrix $C^T$ and synapse onto the recurrent connections of the estimation neurons representing $\mu_{t+1}$, thus implementing the required gradient descent rule. At the same time the estimation neurons project diffuse feedback inhibitory connections back to the sensory prediction errors neurons, representing the $C\mu_{t+1}$ term. 

The dynamical prediction error neurons receive sparse driving excitatory feedforward connections from the estimation neurons, and diffuse inhibitory feedback connections from higher layers encoding the previous state (representing the $A\mu_{t+1}$ term) and, if required, an action efference copy representing the $Bu_t$ term. It has been assumed that the current state is derived not directly from the current estimation, but from a separate processes on a level higher. This is because we assume that if the Kalman filter were implemented in the brain it would not stand alone, but would be deeply integrated within a hierarchical structure that ultimately implements some complex nonlinear filtering scheme. This hierarchical scheme would enable context-dependent Kalman filtering at the lower levels. An additional advantage is that if the Kalman filter only has to deal with local state variables, then it is more likely that the linearity approximations it requires approximately hold. If the system is instead confined to a single layer, so that the estimate at any one time feeds back upon itself, then this can be accomplished with an additional recurrent connection. While this approach could be extended using a nonlinear generative model, and would become a gradient descent version of the extended Kalman filter, the EKF only uses a local linearisation of the nonlinear dynamics and thus performs relatively poorly in highly nonlinear regimes -- thus still necessitating more complex nonlinear representations in higher regions of the brain.

Broadly this scheme accords with predictive coding models of brain \citep*{bastos2012canonical,friston2005theory} and with known general principles of neurophysiology such as diffuse inhibitory feedback connections vs driving feedforward excitatory connections. The only case where this rule does not apply is in the inhibitory feedforward connections between the sensory prediction errors and the estimation units which require precise one-to-one connectivity which is unlikely in the brain. However recent work in predictive coding has shown that such a one-to-one connectivity pattern can empirically be relaxed with relatively little harm to performance \citep{millidge2020relaxing}.

Importantly, this scheme makes adaptive learning of the A,B,and C weight matrices correspond directly to Hebbian plasticity between the prediction error units and the originators of the connection.. For instance, take the A-matrix learning update rule $\frac{dL}{dA} = \Sigma_x \epsilon_x \mu_t^T$. In elementwise terms this can be expressed as: $\frac{dL}{dA_{ij}} = (\Sigma_x \epsilon_x)_i \mu_{t_j}$, which simply makes the update to that connection the product of the variance weighted prediction error and the top-down previous state, which corresponds exactly to a third-factor Hebbian learning rule. 

One potential worry is that implementing the Kalman filter estimation as a gradient descent process rather than an instantaneous inference makes it too slow. In the time it takes the recurrent connections to do multiple gradient descent steps, the world could have moved on so the computed estimation is out of date. In the results below we show, however, that convergence in this case is rapid due to the convexity of the underlying estimation problem. Nearly perfect convergence to the analytical Kalman filter estimate is possible within 5 gradient steps, and reasonable estimates are obtainable within 2 and sometimes even 1 gradient step. Moreover, we found empirically that using too few gradient steps effectively smooths the estimate, potentially a desirable property for the system.
\section{Results}

The proposed model was implemented and tested on a simple Kalman filtering application - that of tracking the motion of an accelerating body given only noise sensor measurements. The body is accelerated with an initial high acceleration that rapidly decays according to an exponential schedule. The Kalman filter must infer the position, velocity, and true acceleration of the body from only a kinematic dynamics model and noisy sensor measurements. The body is additionally perturbed by white Gaussian noise in all of the position, velocity and displacement. The control schedule and the true position, velocity and displacement of the body are shown in figure 1 below.

\begin{figure}[H]
  \begin{subfigure}{0.33\textwidth}
    \centering
    \includegraphics[width=.8\linewidth]{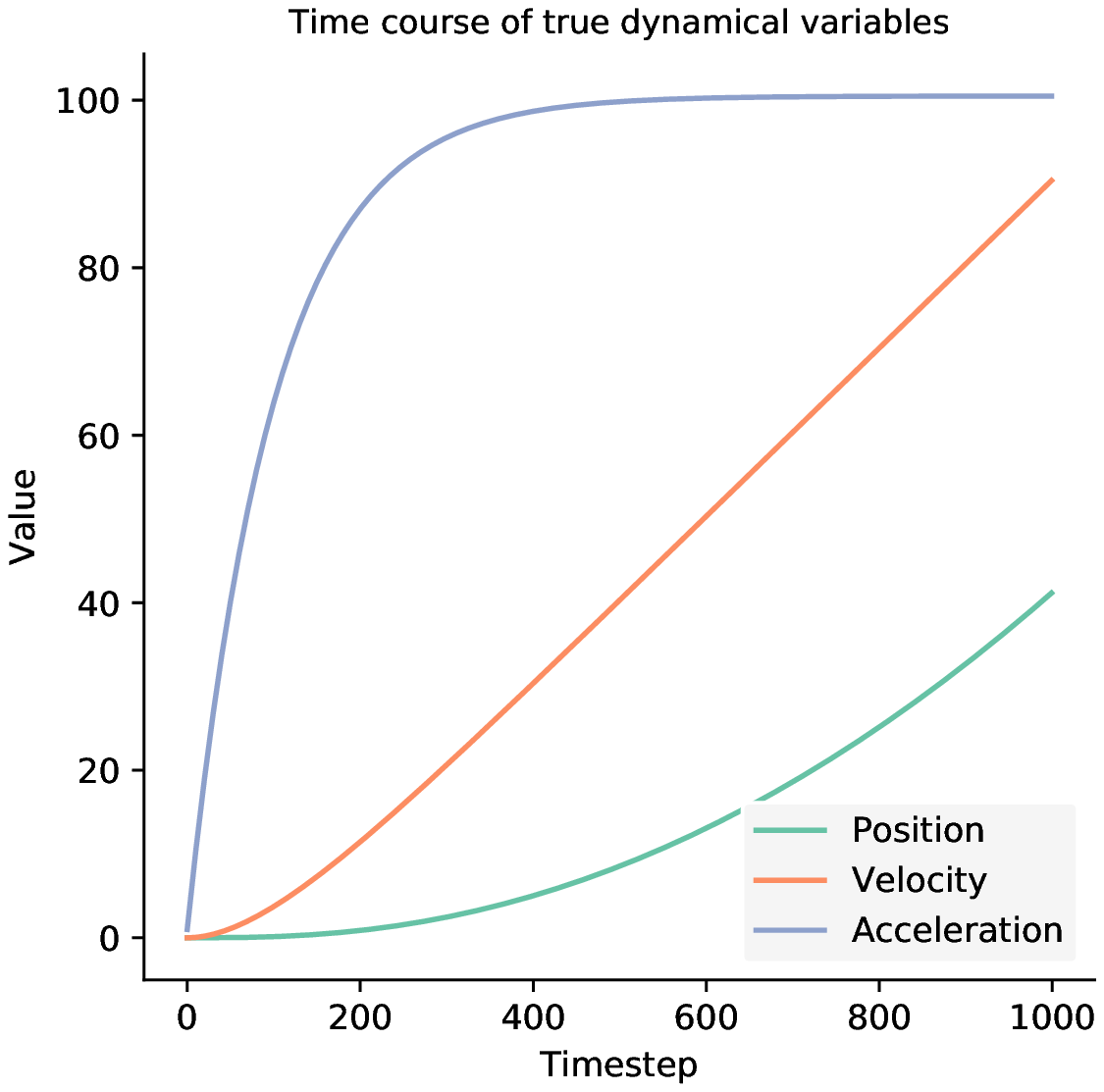}
    \caption{True Dynamics}
  \end{subfigure}%
  \begin{subfigure}{0.33\textwidth}
    \centering
    \includegraphics[width=.8\linewidth]{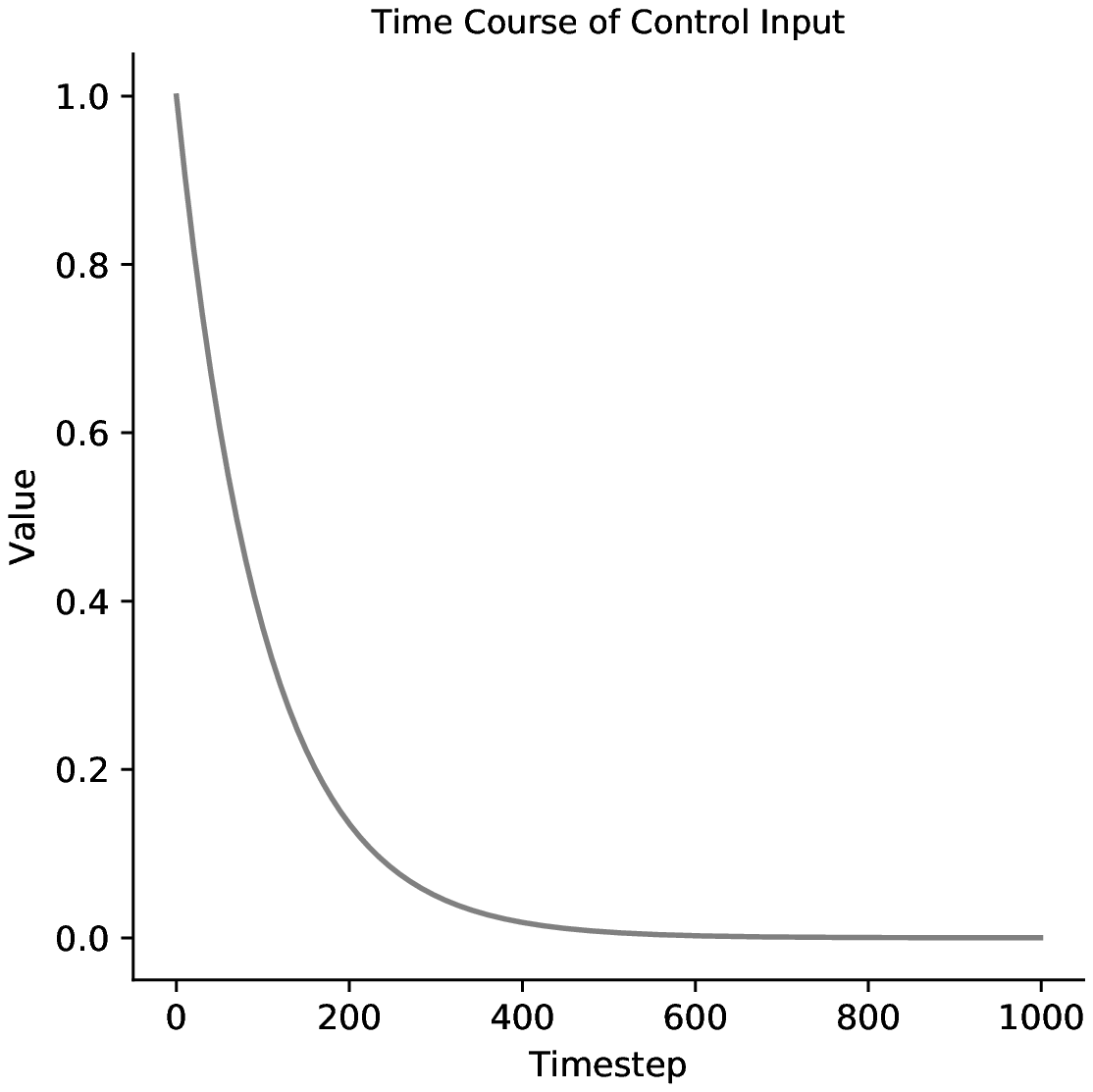}
    \caption{Control Input}
  \end{subfigure}
  \begin{subfigure}{0.33\textwidth}\quad
    \centering
    \includegraphics[width=.8\linewidth]{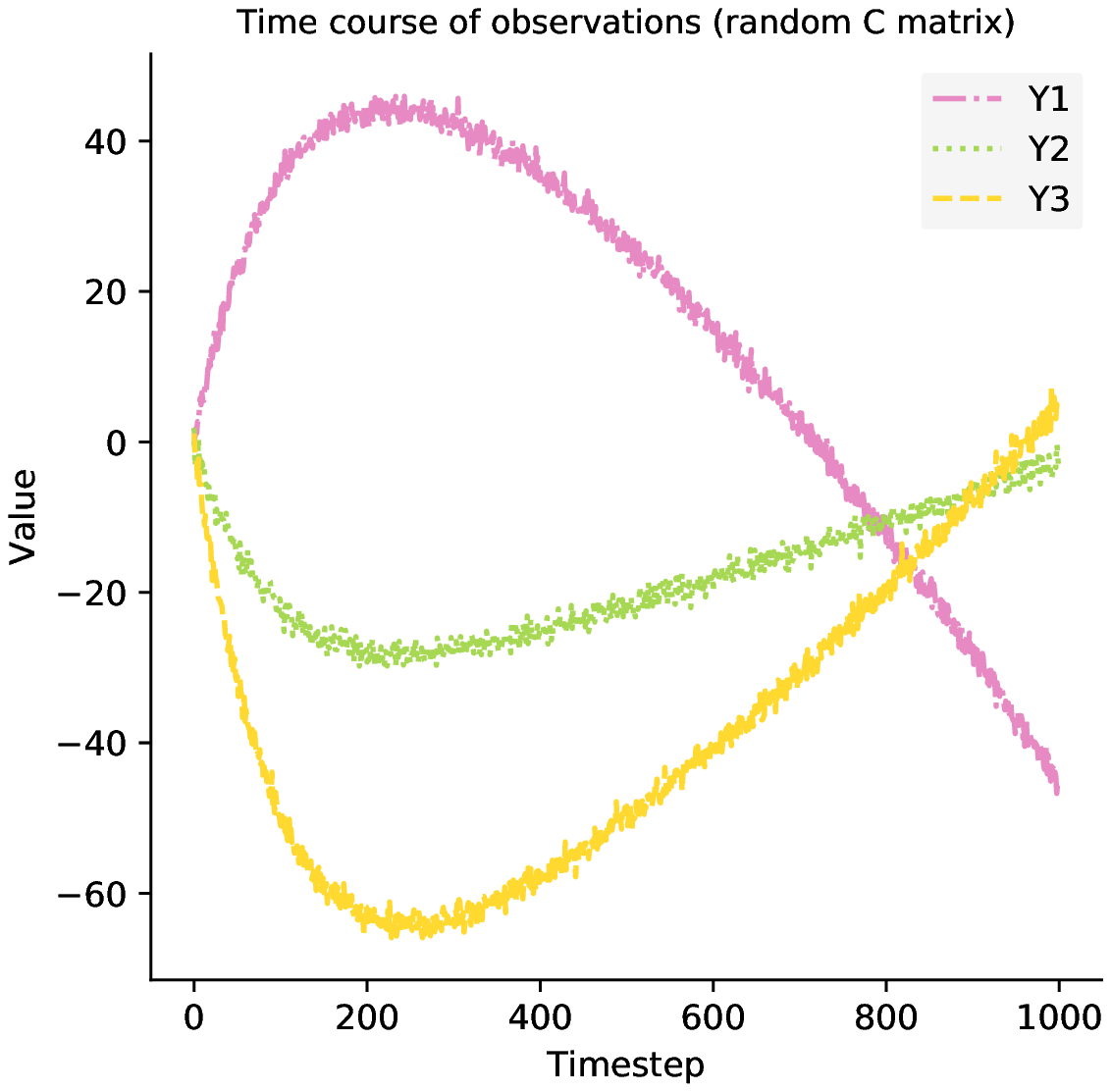}
    \caption{Observations}
  \end{subfigure}
\caption{The true dynamics, control input, and observations generated by a random C matrix}
\end{figure}

The analytical Kalman filter was set up as follows. It was provided with the true kinematic dynamics matrix (A) and the true control matrix (B),
\begin{flalign*}
    A &= \begin{bmatrix}
    1 & dt & \frac{1}{2}dt^2 \\,
    0 & 1 & dt \\
    0 & 0 & 1
    \end{bmatrix} & \\
    B &= \begin{bmatrix}
    0 & 0 & 1
    \end{bmatrix} \numberthis
\end{flalign*}

The observation matrix C matrix was initialized randomly with coefficients drawn from a normal distribution with 0 mean and a variance of 1. This effectively random mapping of sensory states meant that the filter could not simply obtain the correct estimate directly but had to disentangle the measurements first. The Q and R matrices of the analytical kalman filter were set to constant diagonal matrices, where the constant was the standard variance of the noise added to the system.

The performance of the analytical Kalman filter which computed updates using equations 1-3 is compared with that of our neural Kalman filter using gradient descent dynamics.\footnote{The code used for these simulations is freely available and online at $https://github.com/Bmillidgework/NeuralKalmanFiltering$} In this comparison the A, B, and C matrices are fixed to their correct values and only the estimated mean is inferred according to equation 3. Comparisons are provided for a number of different gradient steps. As can be seen in Figure 2, only a small number (5) of gradient descent steps are required to obtain performance very closely matching the analytical result. This is likely due to the convexity of the underlying optimization problem, and means that using gradient descent for "online" inference is not prohibitively slow. The simulation also shows the estimate for too few (2) gradient steps for which results are similar, but the estimate may be slightly smoother.

\begin{figure}[H]
  \begin{subfigure}{0.33\textwidth}
    \centering
    \includegraphics[width=.8\linewidth]{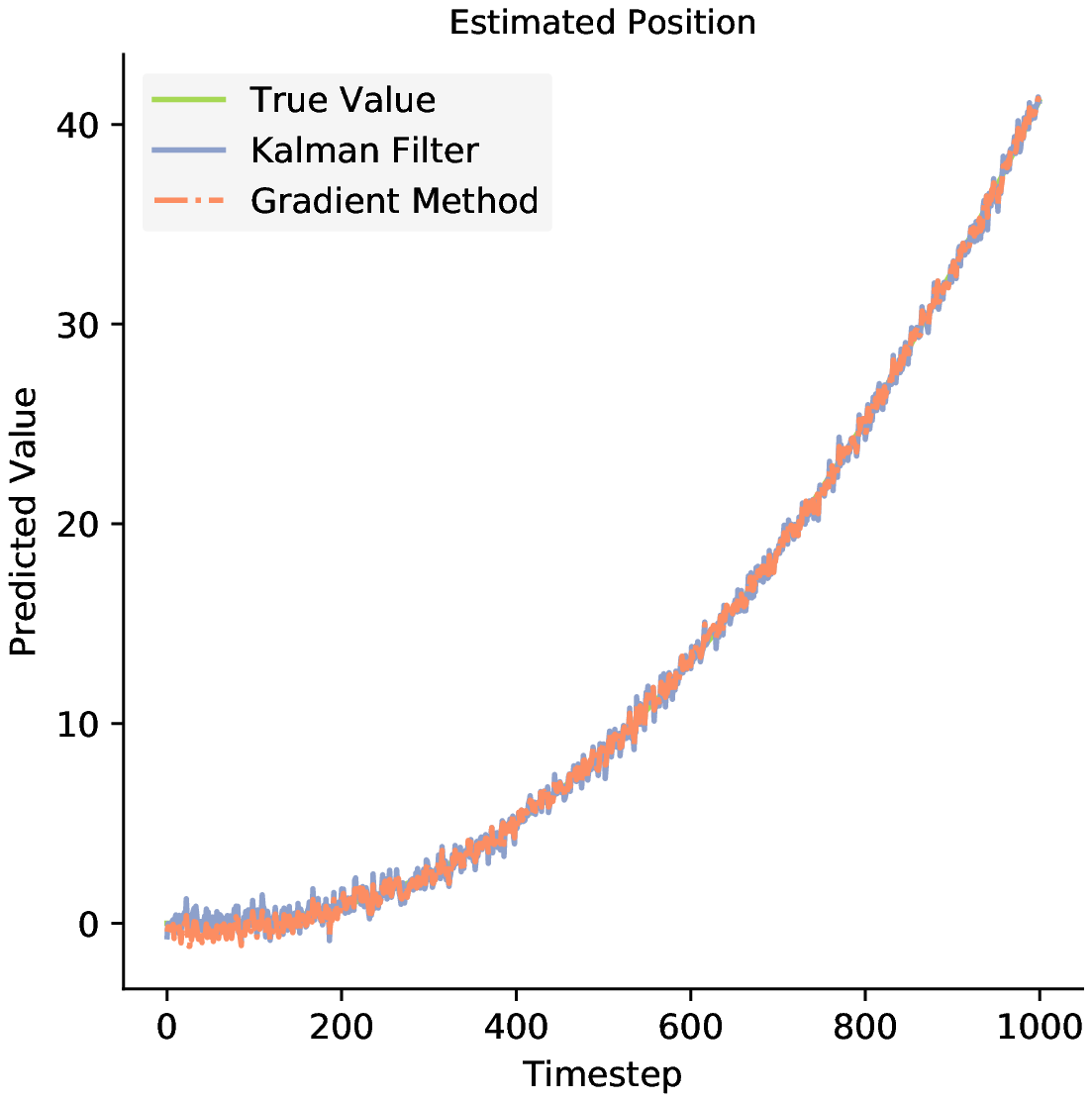}
    \caption{Position}
  \end{subfigure}%
  \begin{subfigure}{0.33\textwidth}
    \centering
    \includegraphics[width=.8\linewidth]{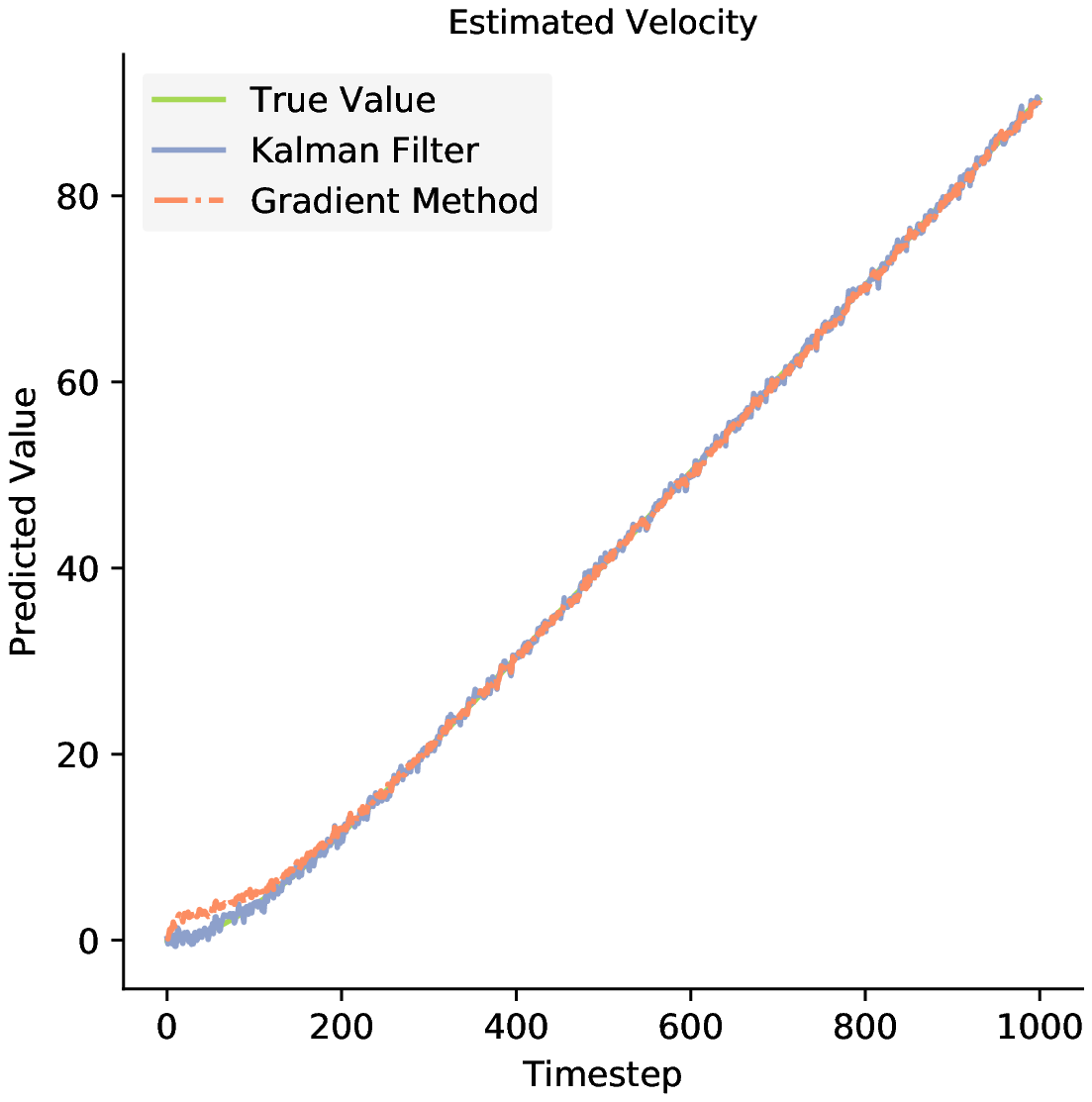}
    \caption{Velocity}
  \end{subfigure}
  \begin{subfigure}{0.33\textwidth}\quad
    \centering
    \includegraphics[width=.8\linewidth]{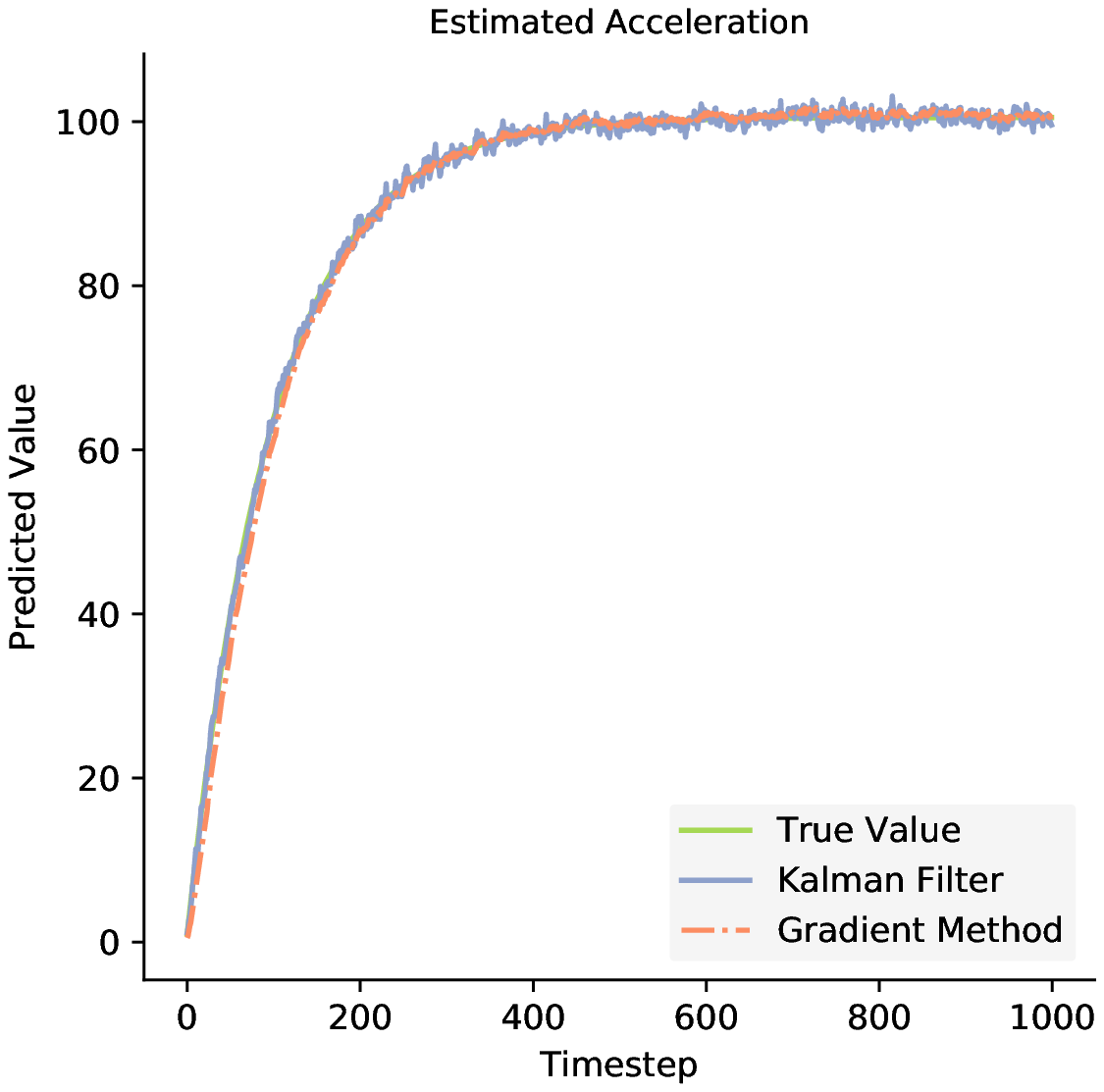}
    \caption{Acceleration}
  \end{subfigure}
  \medskip

  \begin{subfigure}{0.33\textwidth}
    \centering
    \includegraphics[width=.8\linewidth]{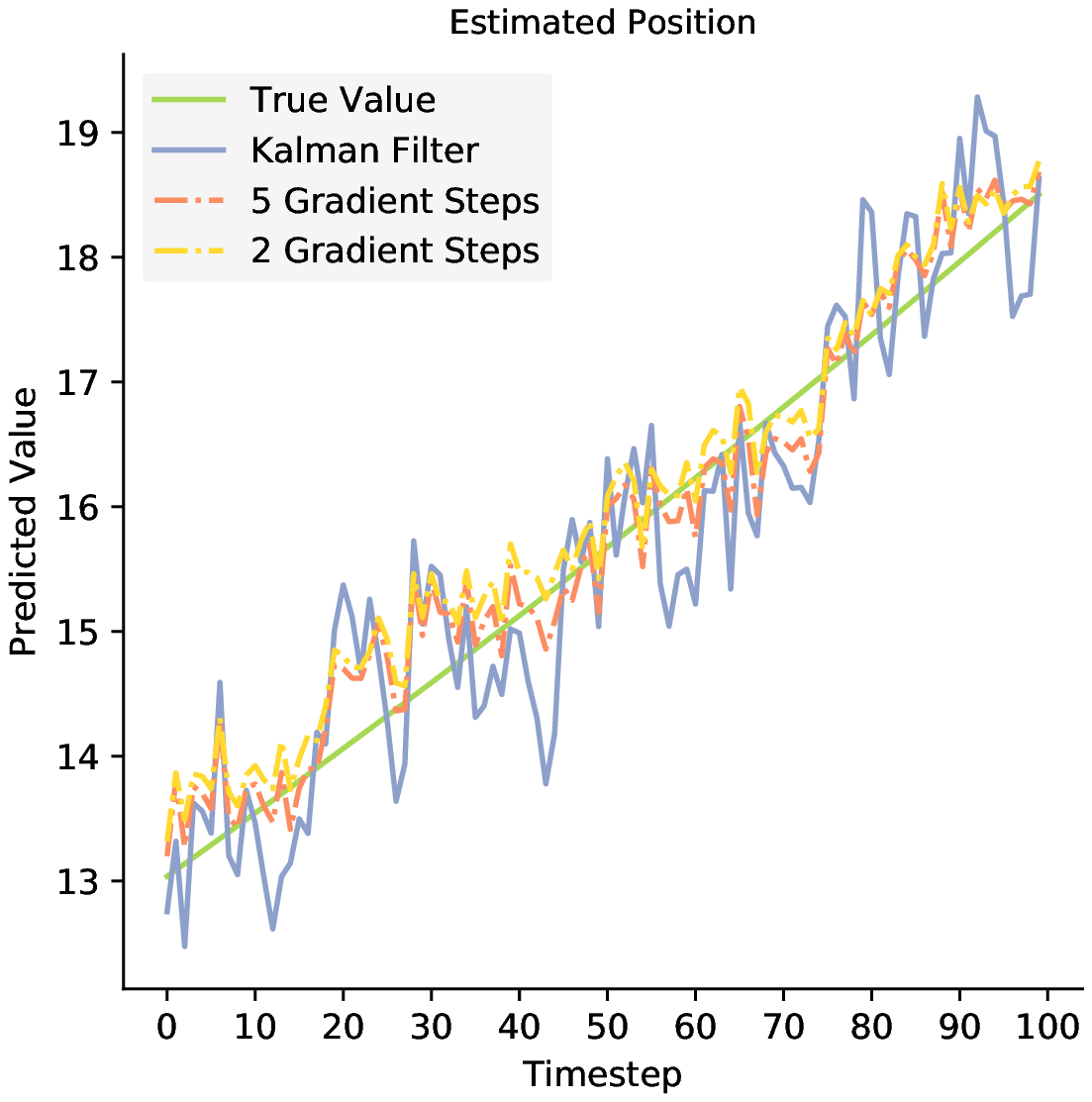}
    \caption{Position}
  \end{subfigure}
  \begin{subfigure}{0.33\textwidth}
    \centering
    \includegraphics[width=.8\linewidth]{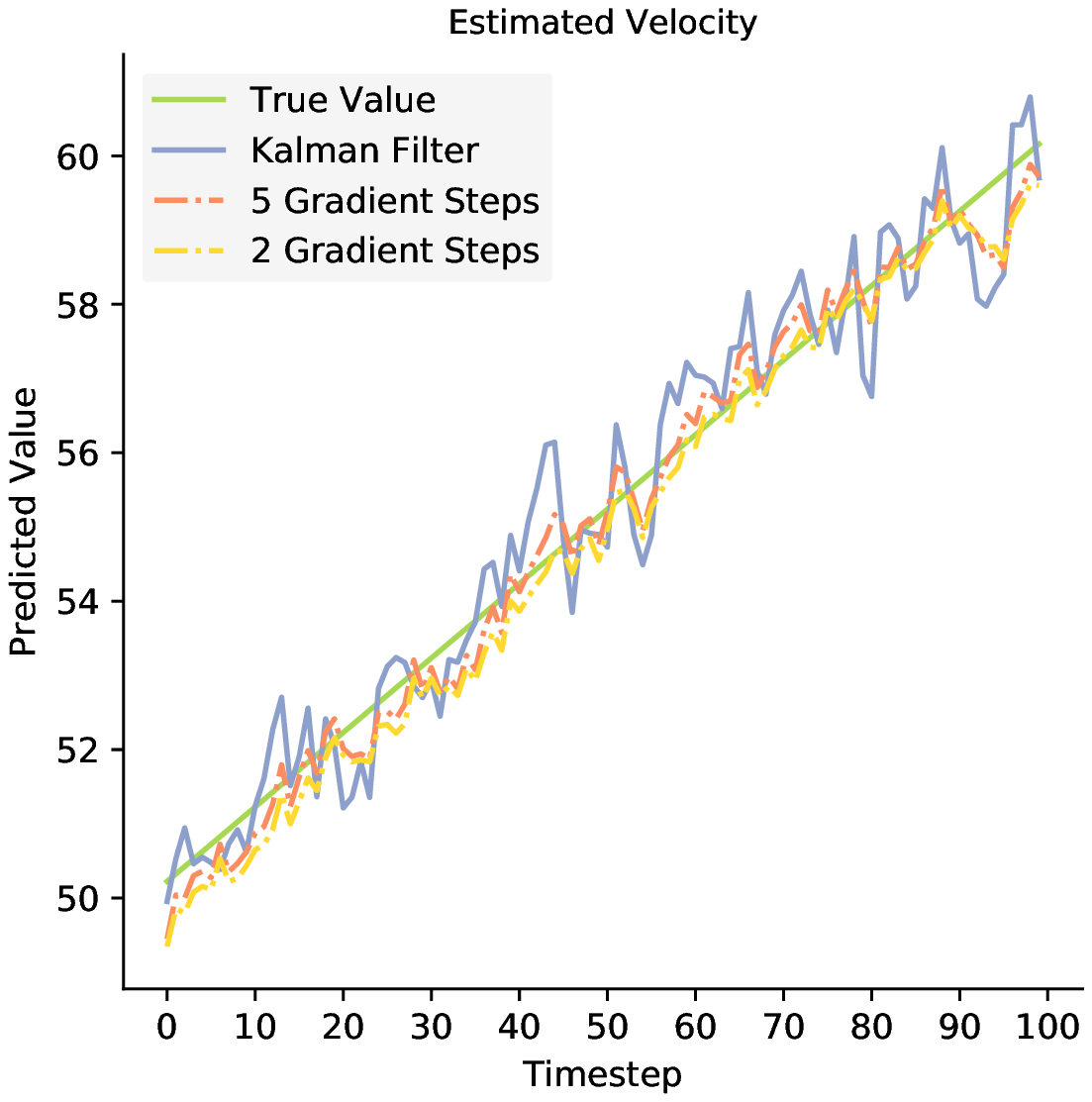}
    \caption{Velocity}
  \end{subfigure}
  \begin{subfigure}{0.33\textwidth}
    \centering
    \includegraphics[width=.8\linewidth]{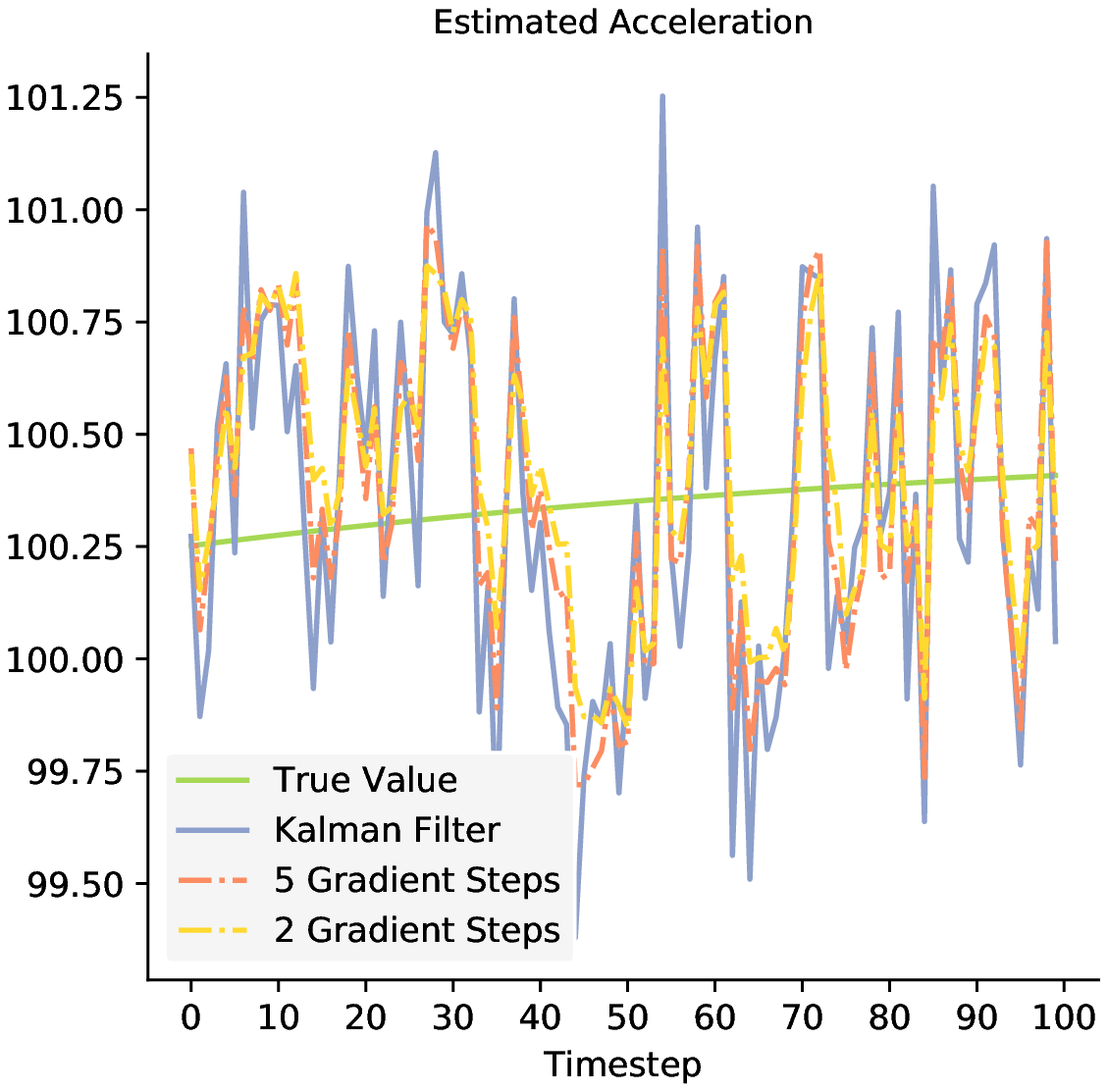}
    \caption{Acceleration}
  \end{subfigure}
  \caption{Tracking performance of our gradient filter compared to the true values and the analytical Kalman Filter. First row shows accurate tracking over 2000 timesteps. Second row zooms in on 100 timestep period to demonstrate tracking performance in miniature and the effect of few gradient updates.}
\end{figure}

Next, we demonstrate the adaptive capabilities of our algorithm. In Figure 3, we show the performance of our algorithm in predicting the position, velocity, and acceleration of the body when provided with a faulty A matrix. Using equation 7, our model learns the A matrix online via gradient descent. To ensure numerical stability, a very small learning rate of $10^{-5}$ must be used. The entries of the A matrix given to the algorithm were initialized as random Gaussian noise with a mean of 0 and a standard deviation of 1. The performance of the algorithm without learning the A matrix is also shown, and estimation performance is completely degraded without the adaptive learning. The adaptivity process converges remarkably quickly. It is interesting, moreover, to compare the matrix coefficients learned through the Hebbian plasticity to the known true coefficients. Often they do not match the true values, and yet the network is able to approximate Kalman filtering almost exactly. Precisely how this works is an area for future exploration. 
If a system similar to this is implemented in the brain, then this could imply that the dynamics model inherent in the synaptic weight matrix should not necessarily be interpretable.

We also show (second row of Figure 3) that, perhaps surprisingly, both the A and B matrix can be learned simultaneously. In the simulations presented below, either only the A matrix, or both the A and the B matrix were initialized with random gaussian coefficients, and the network learned to obtain accurate estimates of the hidden state in these cases. \footnote{The results of only learning the B matrix were extremely similar for that of the A matrix. For conciseness, the results were not included. Interested readers are encouraged to look at the $NKF_AB_matrix.ipynb$ file in the online code where these experiments were run.}

\begin{figure}[H]
  \begin{subfigure}{0.33\textwidth}
    \centering
    \includegraphics[width=.8\linewidth]{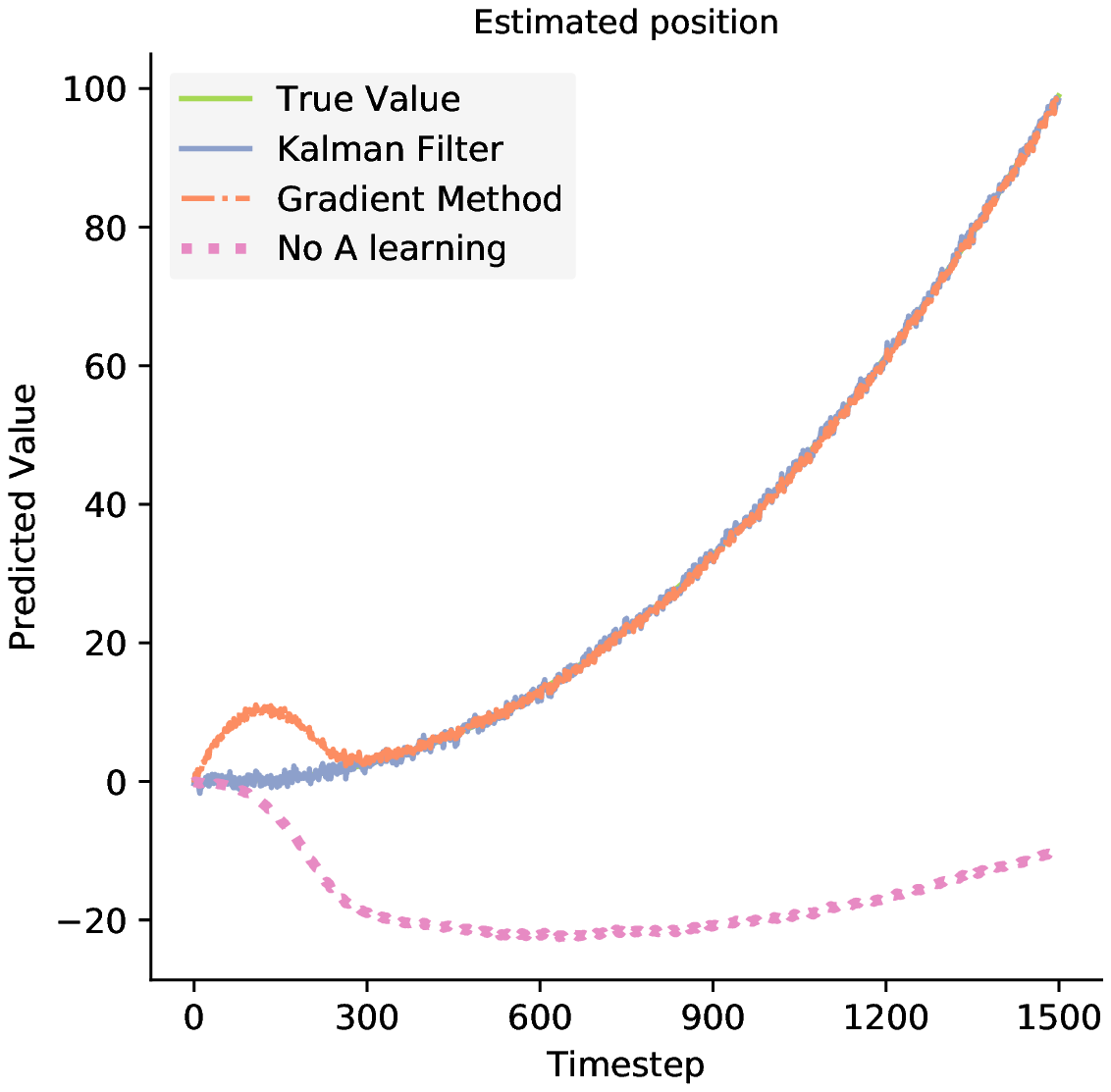}
    \caption{Position for learnt A matrix}
  \end{subfigure}%
  \begin{subfigure}{0.33\textwidth}
    \centering
    \includegraphics[width=.8\linewidth]{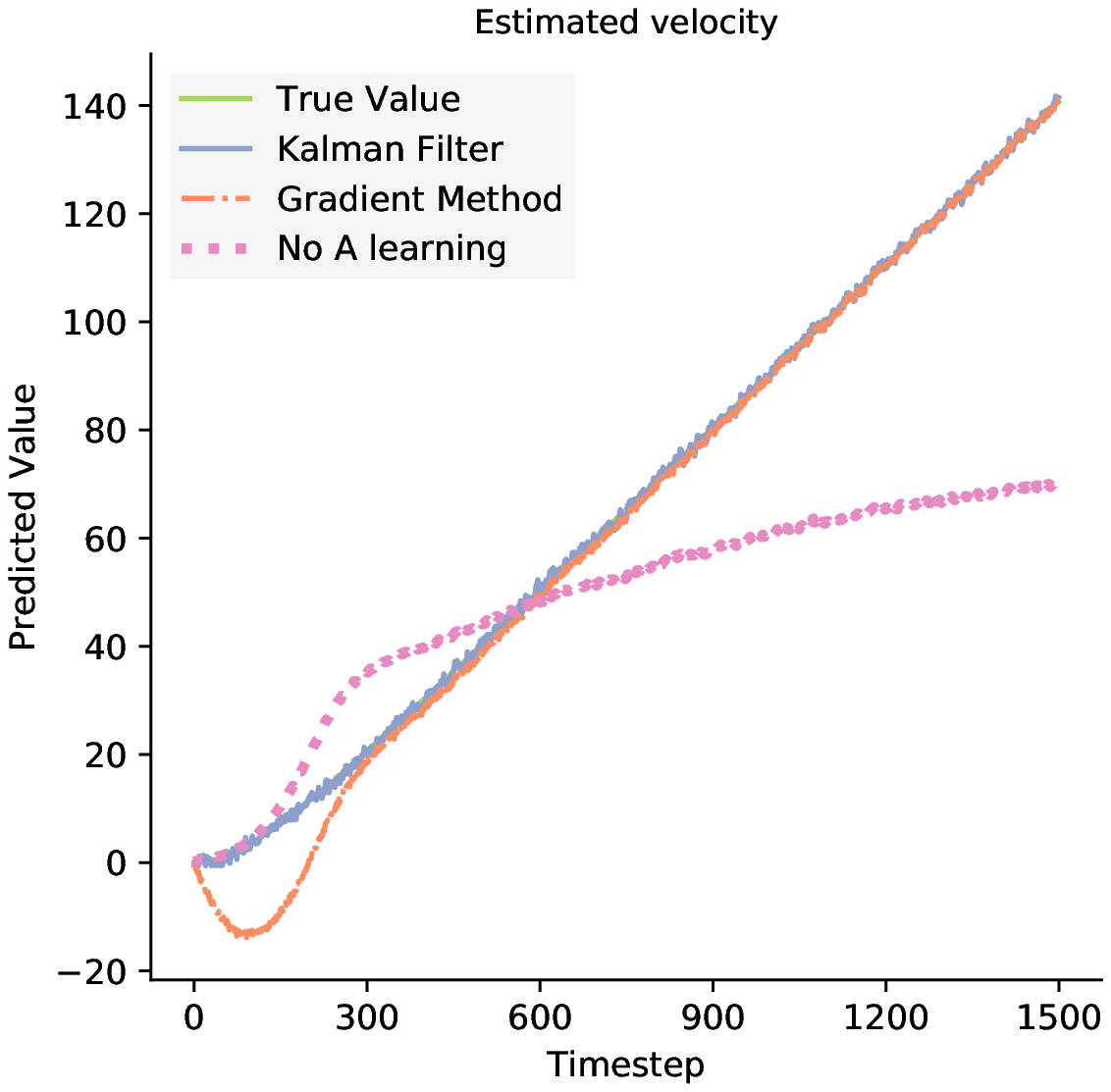}
    \caption{Velocity for learnt A matrix}
  \end{subfigure}
  \begin{subfigure}{0.33\textwidth}\quad
    \centering
    \includegraphics[width=.8\linewidth]{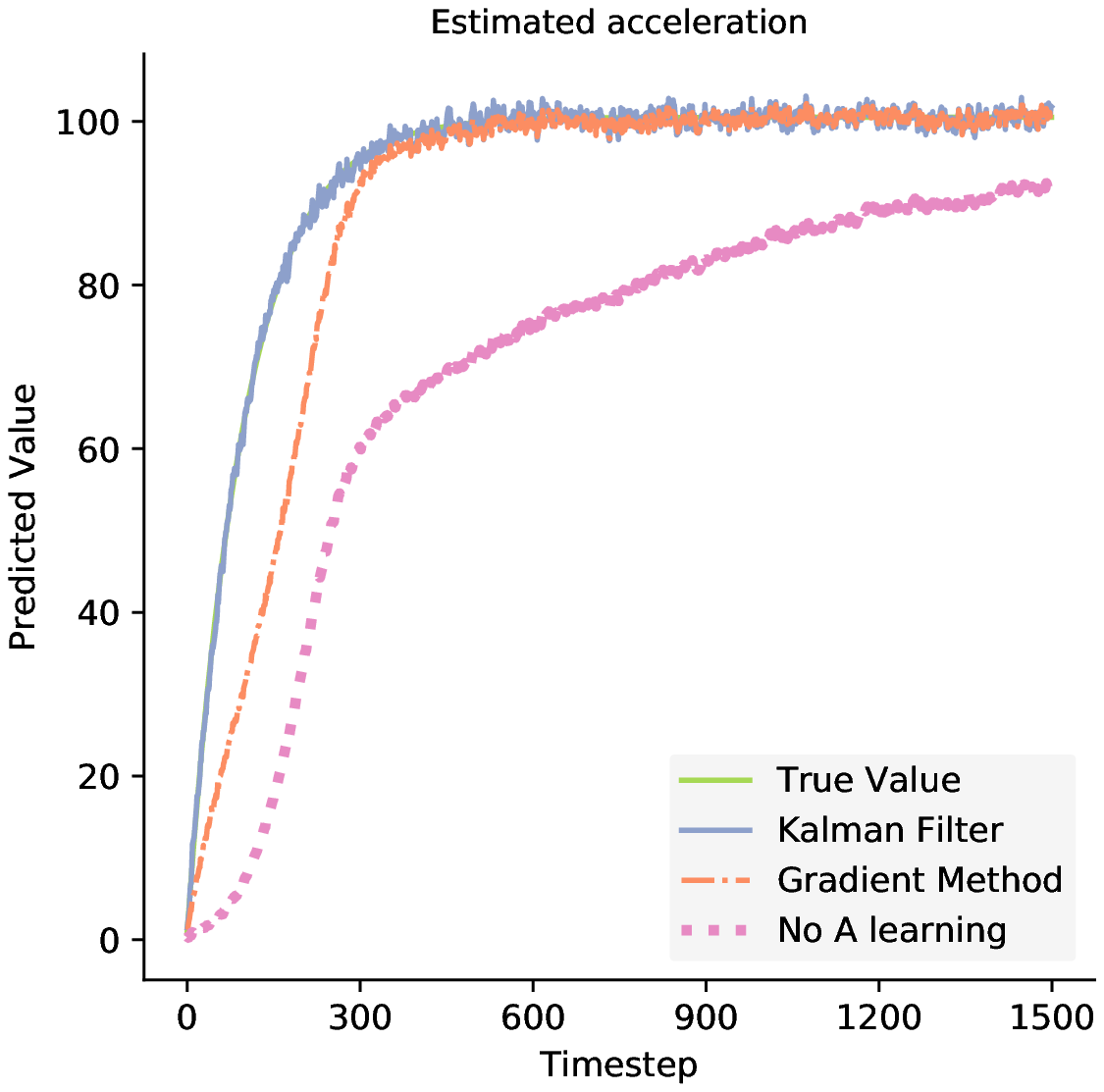}
    \caption{Acceleration for learnt A matrix}
  \end{subfigure}
  \medskip

  \begin{subfigure}{0.33\textwidth}
    \centering
    \includegraphics[width=.8\linewidth]{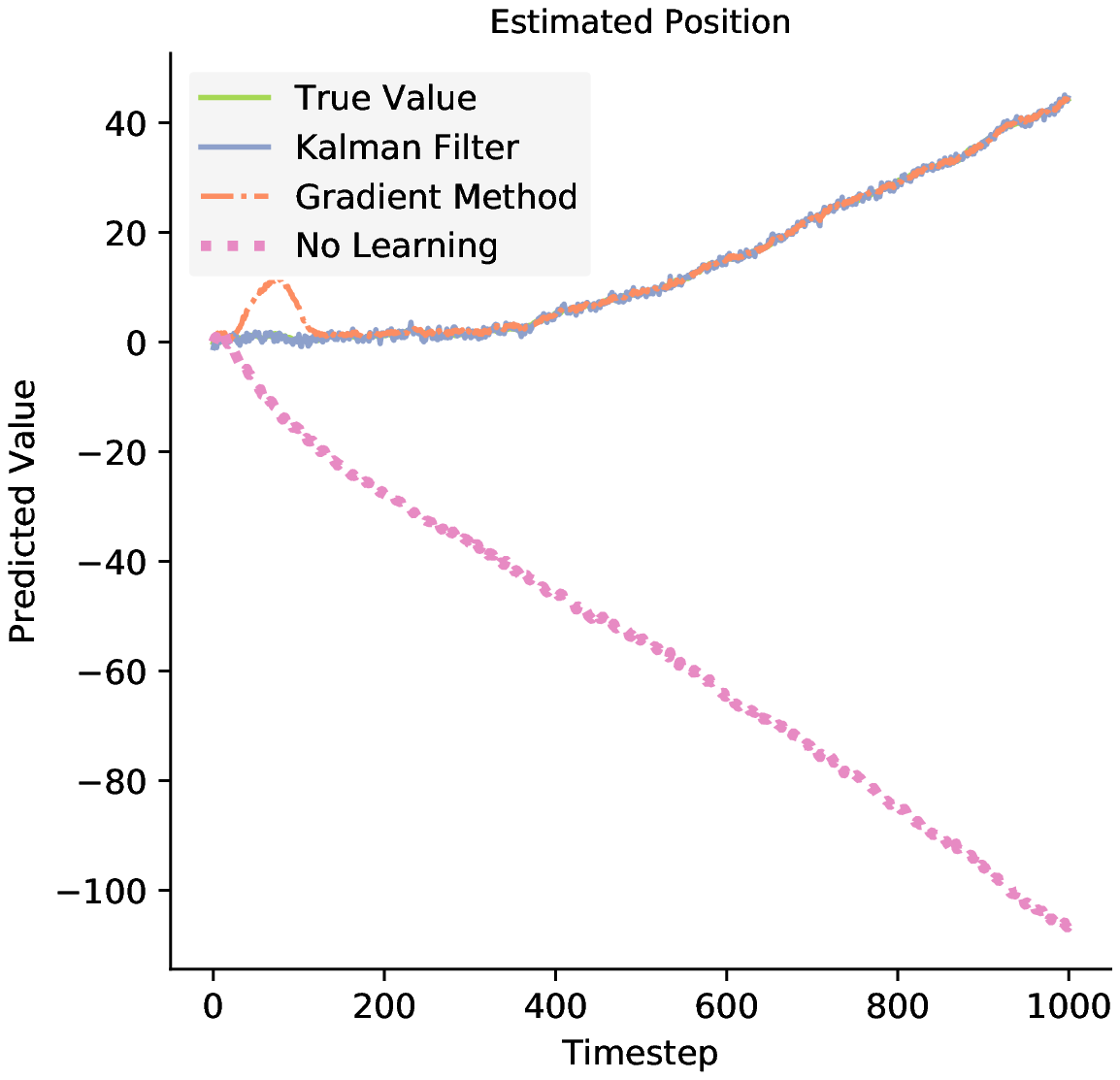}
    \caption{Position: learnt A and B matrices}
  \end{subfigure}
  \begin{subfigure}{0.33\textwidth}
    \centering
    \includegraphics[width=.8\linewidth]{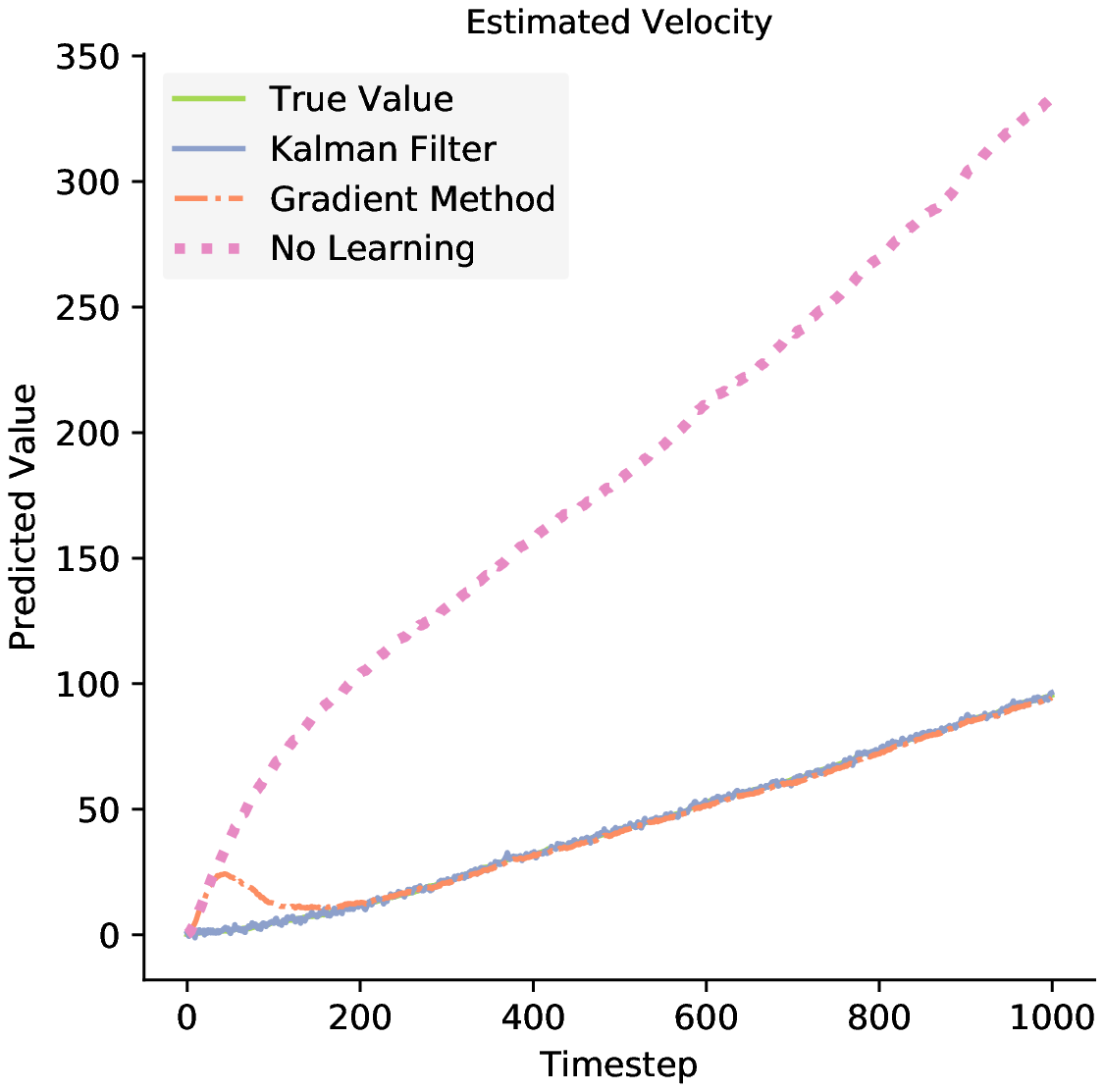}
    \caption{Velocity: learnt A and B matrices}
  \end{subfigure}
  \begin{subfigure}{0.33\textwidth}
    \centering
    \includegraphics[width=.8\linewidth]{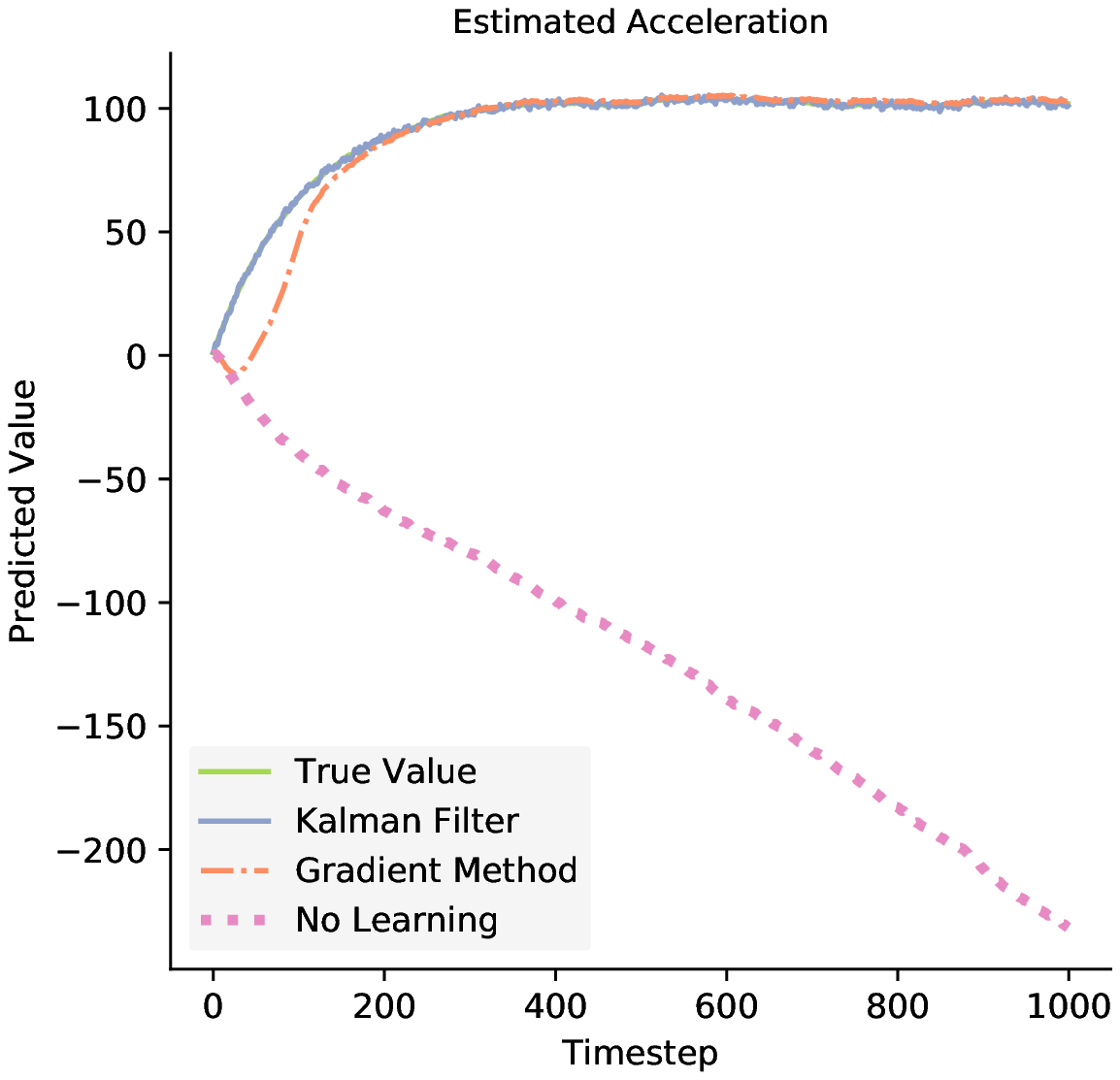}
    \caption{Acceleration: learnt A and B matrices}
  \end{subfigure}
  \caption{Filtering performance for adaptively learning the A matrix (first row) or both the A and B matrices in concert (second row). The filtering behaviour of the of the randomly initialized filters without adaptive learning is also shown.}
\end{figure}

We also tried adaptively learning the C matrix using equation 9, but all attempts to do so failed. Although the exact reason is unclear, we hypothesise that an incorrect C matrix corrupts the observations which provides the only "source of truth" to the system. If the dynamics are completely unknown but observations are known, then the true state of the system must be at least approximately near that implied by the observations, and the dynamics can be inferred from that. On the other hand, if the dynamics are known, but the observation mapping is unknown, then the actual state of the system could be on any of a large number of possible dynamical trajectories, but the exact specifics of which are underspecified. Thus the network learns a C matrix which corresponds to some dynamical trajectory, which succeeds in minimizing the loss function, but which is completely dissimilar to the actual trajectory the system undergoes. This can be seen by plotting the loss obtained according to equation 5 in Figure 4, which rapidly decreases, although the estimate diverges from the true values.

\begin{figure}
  \begin{subfigure}{0.33\textwidth}
    \centering
    \includegraphics[width=.8\linewidth]{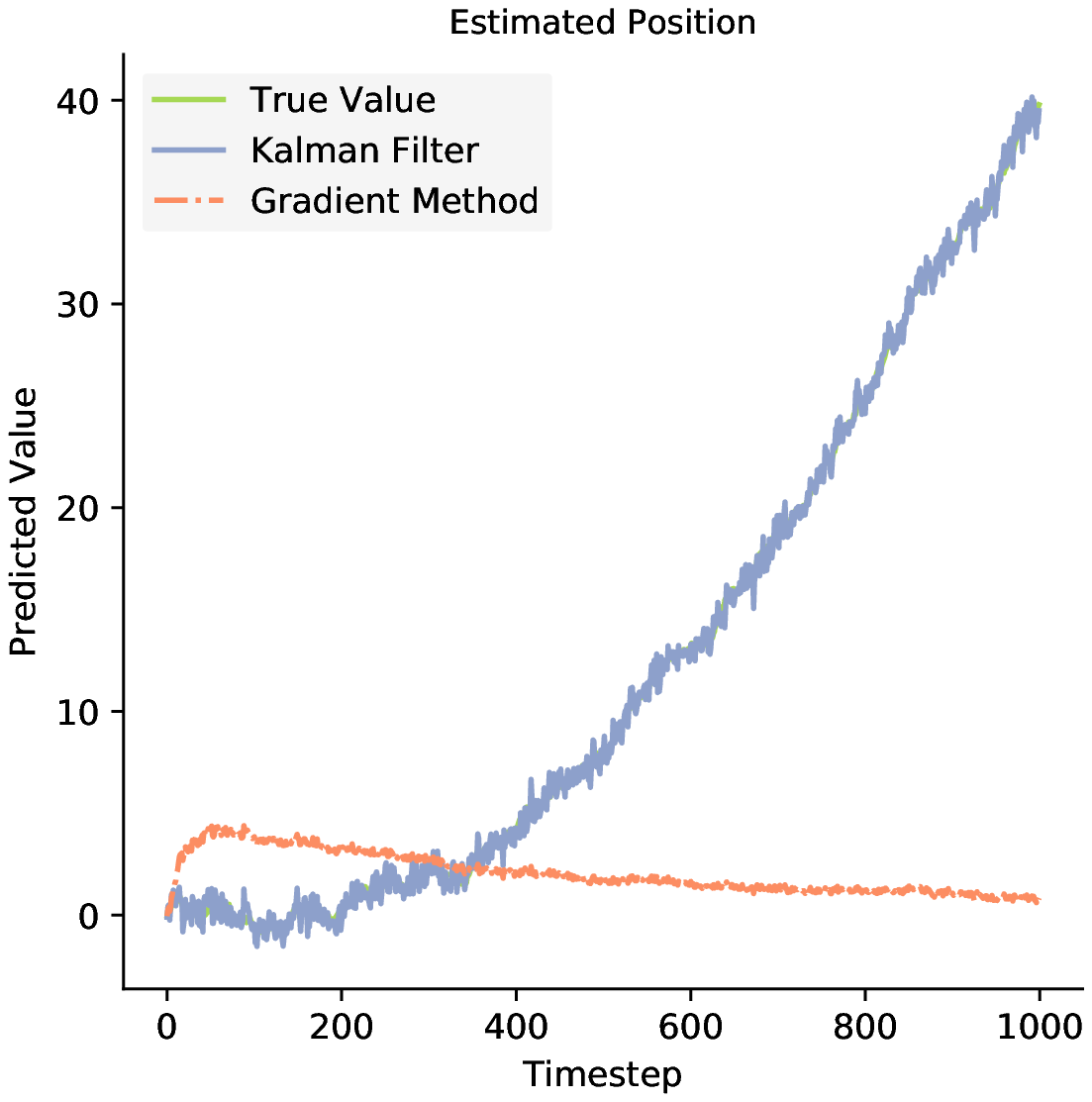}
    \caption{Position: learnt C matrix}
  \end{subfigure}%
  \begin{subfigure}{0.33\textwidth}
    \centering
    \includegraphics[width=.8\linewidth]{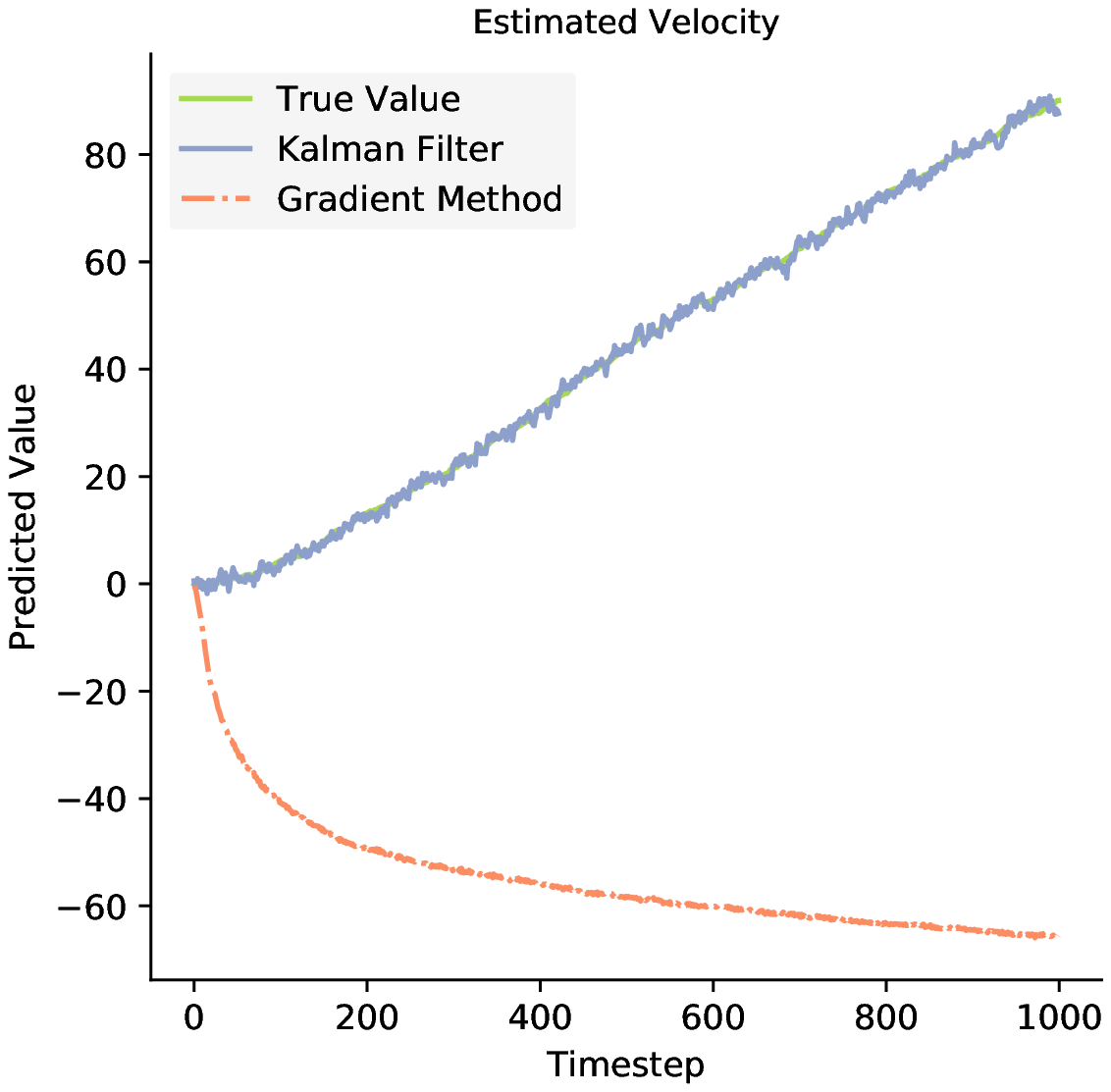}
    \caption{Velocity: learnt C matrix}
  \end{subfigure}
  \begin{subfigure}{0.33\textwidth}\quad
    \centering
    \includegraphics[width=.8\linewidth]{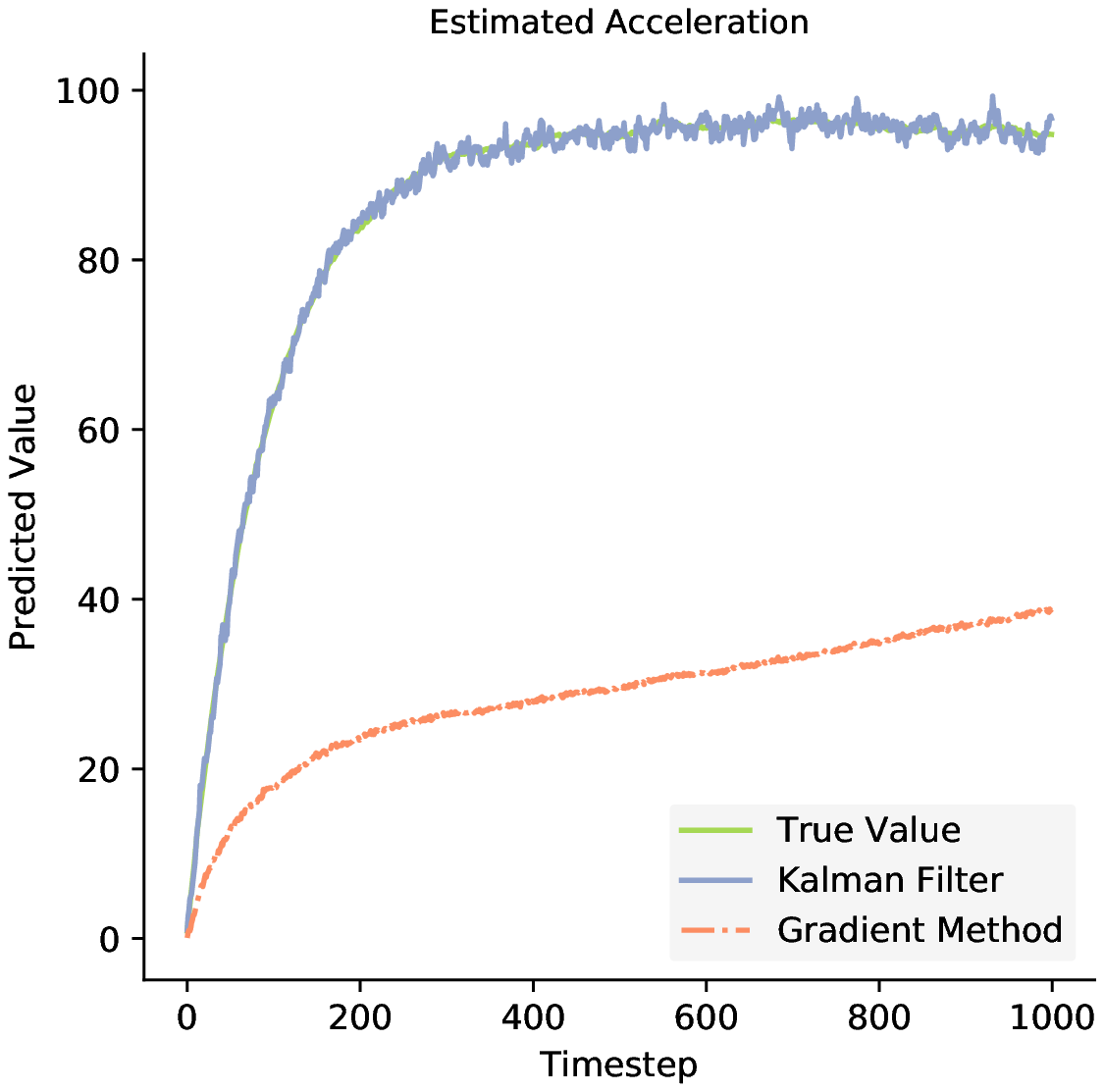}
    \caption{Acceleration: learnt C matrix}
  \end{subfigure}
  \medskip

  \begin{subfigure}{0.5\textwidth}
    \centering
    \includegraphics[width=.8\linewidth]{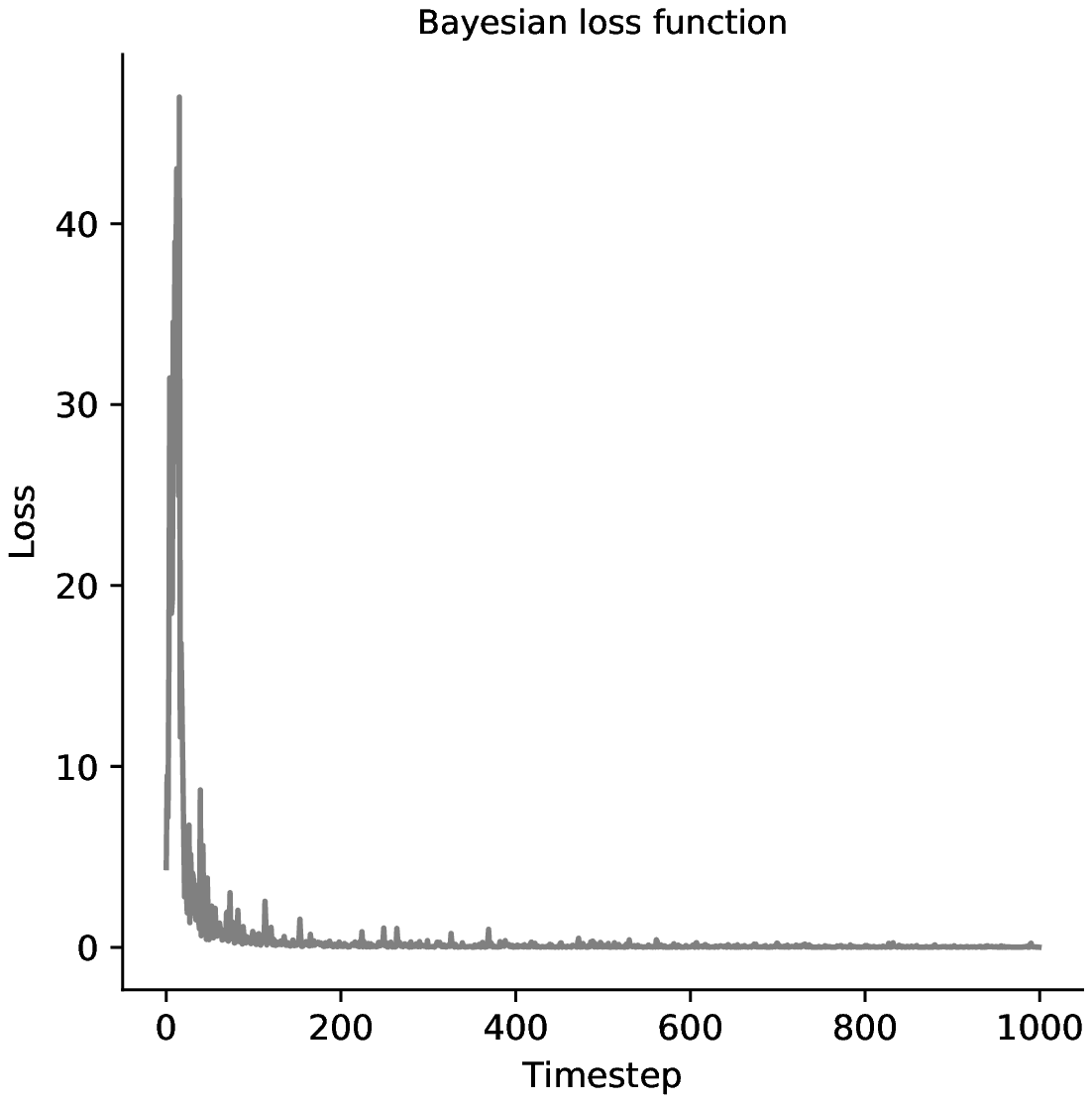}
    \caption{Loss function over timestels}
  \end{subfigure}
  
  \caption{Very poor tracking behaviour with a learnt C matrix. This is despite the fact that the bayesian loss function rapidly decreases to a minimum. This shows that the filter can find a prediction-error minimizing "solution" which almost arbitrarily departs from reality if the C-matrix is randomized.}
\end{figure}

\section{Discussion}
Thus far we have derived a gradient descent approximation to the Kalman filtering equations. We have shown that the estimated state converges rapidly to the exact analytical result. The gradients imply an update rule which is a simple sum of variance weighted prediction errors, and which can be straightforwardly computed in a network of rate-coded integrate-and-fire neurons.

We have also shown that our gradient interpretation allows one to derive simple rules for learning the dynamics matrices A and B, which correspond to Hebbian plasticity, and shown empirically that these rules can be used to obtain accurate filtering performance even when the initial dynamics matrices provided to the system are Gaussian noise. 

Due to only requiring prediction errors and Hebbian plasticity, we have claimed that our model is biologically plausible. We here interpret biological plausibility to require local and simple (linear) computation. This means that the computation occurring at each "neuron" can only require information that could in principle be available at that neuron or synapse. Additionally, we have assumed that neurons can only do simple operations on inputs - primarily summation of incoming weighted inputs. We have also assumed that synaptic weights are multiplicative.

However, there are several deficiencies with our model in terms of biological plausibility which it is important to state. Our model assumes full connectivity for the "diffuse" connectivity required to implement matrix multiplications. Additionally in other cases it requires one-to-one excitatory connectivity, both constraints which are not fully upheld in neural circuitry. Additionally, in one case (that of the "C matrix" between the populations of neurons representing the estimate and the sensory prediction errors), we have assumed a complete symmetry of backward and forward weights, such that the connections which embody the C matrix downwards also implement the $C^T$ matrix when traversing upwards. This is also a constraint not satisfied within the brain. Additionally, our model can represent negative numbers in states or prediction errors, which rate-coded neurons cannot. The robustness of our method to the tightening of these constraints is thus an interesting topic for future investigation.

We believe, however, that despite some lack of biological plausibility, our model is useful in that it shows how a standard engineering algorithm can be derived in a way more amenable to neural computation, and provides a sketch at how it could be implemented in the brain. Moreover, we hope to draw attention to Bayesian filtering algorithms and how they can be implemented neurally, instead of just Bayesian inference on static posteriors.

While our implementation uses the prior state derived from the previous output of the algorithm, it is also possible that in the brain the prior state could be set by, or influenced by feedback connections from higher levels, representing more advanced states of processing which could inform the state estimate. 
We believe this is an important point about how Kalman filtering may fit into the larger picture. Since Kalman filtering turns out to be relatively straightforward algorithm which can be implemented in a rather small neural circuit, it could perhaps serve as a composable building block of cortical processing. Kalman filters could potentially be implemented at the lowest hierarchical levels to achieve some immediate processing and reduction of sensory noise before passing on improved state estimates to higher levels of processing which can then apply more complex nonlinear filtering algorithms. Using a Kalman filter locally, at the bottom of the hierarchy, would also reduce the primary disadvantage of the Kalman filter: its assumption of linearity. If restricted to dealing with local state and observations, the linearity approximation would become substantially more accurate, and the Kalman filter could provide valuable noise reduction and sensor-fusion properties to state estimators at higher levels. While the higher levels would provide more global contextualization of the prior state at each filtering step.

\section{Conclusion}

We have shown that it is possible to derive a gradient descent approximation to the Kalman filter which requires only local computation and simple operations such as the summation of variance weighted prediction errors. We have proposed a simple neural architecture of rate-coded integrate-and-fire neurons which can implement this algorithm. We have additionally shown that the gradient descent approach also provides a learning rule for adaptively learning the dynamics model which is identical to Hebbian plasticity. We have shown that our algorithm rapidly converges to the estimate of the exact Kalman filter, and that the Hebbian plasticity rule we have derived enables dynamics models to be learnt online during filtering.

\bibliography{cites.bib}

\section{Appendix A: Derivation of Kalman Filtering Equations from Bayes' Rule}

In this appendix we derive the Kalman filtering equations directly from Bayes rule. The first step is to derive the projected covariance,
\begin{align}
    \E[\hat{x}_{t+1}\hat{x}_{t+1}^T] &= \E[(Ax + Bu + \omega)(Ax + Bu + \omega)^T] \\
    &= \E[Ax x^TA^T] + \E[Ax u^TB^T] + \E[Ax\omega^T] + \E[Bu x^TA^T] + \E[Bu\omega^T] + \E[\omega_x^T A^T] + \E[\omega U^TB^T] + \E[\omega\omega^T] \\
    &= A \E[xx^T]A^T + \E[\omega\omega^T] \\
    &= A\Sigma_x(t)A^T + \Sigma_\omega 
\end{align}
Step 12 uses the fact that matrices A,B are constant so come out of the expectation operator, and that it is assumed that covariances between the state, the noise, and the control -- $\E[x u^T]$,$\E[x\omega^T]$, $\E[u\omega^T]$ -- are 0. Step 14 uses the fact that $\E[xx^T] = \Sigma_x(t)$ and that $\E[\omega\omega^T] = \Sigma_\omega$.

Next we optimize the following loss function, derived from Bayes' rule above (Equation 5).
\begin{flalign}
     L &= -(y - C\mu_{t+1})^T\Sigma_Z(y - C\mu_{t+1}) + (\mu_{t+1} - A\mu_t - Bu_t)^T\hat{\Sigma}_x(\mu_{t+1} - A\mu_t - Bu_t) &
\end{flalign}
To obtain the Kalman estimate for $\mu_{t+1}$ we simply take derivatives of the loss, set it to zero and solve analytically.
\begin{flalign}
    0 &= \frac{dL}{d\mu_{t+1}}[\mu_{t+1}^T[C^T RC + \Sigma_x]\mu_{t+1} - \mu_{t+1}^T[C^T Ry - \Sigma_x A\mu_t - \Sigma_x Bu_t] - [y^TR - \mu_t^TA^T\Sigma_x - u_t^TB^T \Sigma_x]\mu_{t+1} \\
    &= 2[C^T RC + \Sigma_x]\mu_{t+1} - 2[C^T Ry + \Sigma_x (A \mu_t + Bu_t)] \\
    \mu_{t+1} &= [C^T RC + \Sigma_x^{-1}[C^T Ry + \Sigma_x (A \mu_t + Bu_t] \\
    &= [\Sigma_x^{-1} - \Sigma_x^{-1}C^T[C\Sigma_x C^T + R]^{-1}C\Sigma_x^{-1}[C^T Ry + \Sigma_x (A \mu_t + Bu_t)] \\
    &= [\Sigma_x^{-1} - KC\Sigma_x^{-1}][C^T Ry + \Sigma_x (A \mu_t + Bu_t)] \\
    &= A\mu_t + Bu_t + \Sigma_x^{-1}C^T Ry - KC\Sigma_x^{-1}C^T Ry - KC(A\mu_t + Bu_t) \\
    &= \hat{\mu_{t+1}} - KC\hat{\mu_{t+1}} + [\Sigma_x^{-1}C^TR - KC\Sigma_x^{-1}C^TR]y \\
    &= \hat{\mu_{t+1}} - KC\hat{\mu_{t+1}} + K K^{-1}[\Sigma_x^{-1}C^TR - KC\Sigma_x^{-1}C^TR]y \\
    &= \hat{\mu_{t+1}} - KC\hat{\mu_{t+1}} + K[(C\Sigma_x C^T + R)C^{-T}\Sigma_x[\Sigma_x^{-1}C^TR] - C\Sigma_x C^T R]y \\
    &= \hat{\mu_{t+1}} - KC\hat{\mu_{t+1}} + K[(C\Sigma_x C^T + R^{-1})R - C\Sigma_x C^T R]y \\
    &= \hat{\mu_{t+1}} - KC\hat{\mu_{t+1}} + Ky \\
    &= \hat{\mu_{t+1}} + K[y-C\hat{\mu_{t+1}}]
\end{flalign}
Where $K =\Sigma_x^{-1}C^T[C \Sigma_x C^T + R]^{-1}$ and is the Kalman gain and $\hat{\mu_{t+1}} = A \mu_t + Bu_t$ is the projected mean. 

The first few steps rearrange the loss function into a convenient form and then derive an expression for $\mu_{t+1}$ directly. Step 22 applies the Woodbury matrix inversion lemma to the $[C^TRC + \Sigma_x]^{-1}$ term. The next step rewrites the formula in terms of the Kalman gain matrix K and  multiplies it through. The other major manipulation is the multiplication of the last term of equation 23 by $KK^{-1}$ which is valid since $KK^{-1} = I$.

This derives the optimal posterior mean as the analytical solution to the optimization problem. Deriving the optimal covariance is straightforward and done as follows,

\begin{align}
    \E[\mu_{t+1} \mu_{t+1}^T] &= \E[(\hat{mu_{t+1}} + Ky - KC\hat{mu_{t+1}})(\hat{\mu_{t+1}} + Ky - KC\hat{\mu_{t+1}})^T] \\
    &= \E[\hat{\mu_{t+1}}\hat{\mu_{t+1}}^T] - \E[\hat{\mu_{t+1}}\hat{\mu_{t+1}}^T]C^T K^T - K C \E[\hat{\mu_{t+1}}\hat{\mu_{t+1}}^T] + K \E[y y^T]K^T + K C \E[\hat{\mu_{t+1}}\hat{\mu_{t+1}}^T]C^T K^T \\
    &= \Sigma_{\hat{\mu_{t+1}}} - \Sigma_{\hat{\mu_{t+1}}} C^T K^T - KC \Sigma_{\hat{\mu_{t+1}}} + K[R + C\Sigma_{\hat{\mu_{t+1}}}C^T]K^T \\
   &=  \Sigma_{\hat{\mu_{t+1}}} - \Sigma_{\hat{\mu_{t+1}}} C^T K^T - KC \Sigma_{\hat{\mu_{t+1}}} + \Sigma_{\hat{\mu_{t+1}}}C^T[C\Sigma_{\hat{\mu_{t+1}}}C^T +R]^{-1}[R + C\Sigma_{\hat{\mu_{t+1}}}C^T]K^T \\
   &= \Sigma_{\hat{\mu_{t+1}}} - \Sigma_{\hat{\mu_{t+1}}} C^T K^T - KC \Sigma_{\hat{\mu_{t+1}}} +  \Sigma_{\hat{\mu_{t+1}}} C^T K^T \\
   &= \Sigma_{\hat{\mu_{t+1}}} - KC \Sigma_{\hat{\mu_{t+1}}} \\
   &= [I - KC]\Sigma_{\hat{\mu_{t+1}}}
\end{align}

Which is the Kalman update equation for the optimal variance. The second line follows on the assumption that $\E[x y^T] = 0$. On equation 32 the definition of the Kalman gain is substituted back in and the two $C\Sigma_x C^T + R$ terms cancel.

\section{Appendix B: Relationship between Predictive Coding and Kalman Filtering}

In this appendix we clarify the relationship between predictive coding and Kalman filtering. To jump ahead, we demonstrate that predictive coding and Kalman filtering optimize the same objective -- effectively performing maximum-a-posteriori inference over a trajectory of states and observations -- however predictive coding optimizes the objective via gradient descent on the mean state variable $\mu_t$ while Kalman filtering instead solves this (convex) optimization problem analytically resulting in an exact single-step update. Due to its exact nature, for engineering applications the Kalman filter solution should be preferred, however the gradient descent equations of predictive coding are simple, elegant, and local in nature, allowing them to be performed efficiently, in theory, by distributed neural circuitry.

Predictive coding is usually derived as a variational inference algorithm. Variational inference is a class of approximate Bayesian inference methods which are used to turn an inference problem of computing a posterior directly, into an inference problem of optimizing the parameters $\theta$ of a separate variational distribution $q(x;\theta)$ to be as close as possible to the true posterior. In most cases, this algorithm is approximate as the variational distribution is chosen to be from a simpler family of distributions than the true posterior so as to allow for efficient computation. However, in this case the variational family is of the same family as the true posterior (both are Gaussian) and so usually approximate variational inference methods become exact.

Variational inference proceeds by defining a variational distribution $q(x;\theta)$ which is to be optimized and minimizing the divergence between this distribution and the true posterior. Since the true posterior is assumed to be intractable (which is why we would use variational inference in the first place), we cannot directly minimize the divergence between them, but instead minimize the tractable variational bound, which is called the variational free energy $\mathcal{F}$ \footnote{In the literature the negative of this bound is also called the Evidence Lower Bound (ELBO)}.
\begin{align*}
    \mathcal{F}(y) &= \KL[q(x;\theta) || p(y,x)] \\
    &= \KL[q(x;\theta)||p(x |y)] - \ln p(y) \\
    &\geq \KL[q(x;\theta)||p(x |y)] \numberthis
\end{align*}
To evaluate $\mathcal{F}$, it is necessary to specify the variational density $q(x;\theta)$ and the generative model $p(y,x)$. In the case of Kalman filtering, we are interested in performing variational inference over full trajectories $y_{1:T}, x_{1:T}$ of observations and hidden states. Importantly, it is straightforward to demonstrate that with a Markov factorization of the generative model $p(y_{1:T}, x_{1:T}) = p(y_1 | x_1)p(x_1) \prod_{t=2}^T p(y_t | x_t)p(x_t | x_{t-1})$ \footnote{Where, to make this expression not a function of $x_{t-1}$, we implicitly average over our estimate of $x_{t-1}$ from the previous timestep: $p(x_t | x_{t-1}) = \E_{q(x_{t-1})}[p(x_t | x_{t-1})]$} and a mean-field temporal factorization of the variational density, so that it is independent across timesteps $q(x_{1:T} ; \theta) = \prod_{t=1}^T q(x_t ; \theta)$, then the free energy of the trajectory factorizes into independently optimizable free-energies of a particular timestep,
\begin{align*}
    \mathcal{F}(y_{1:T}) &= \sum_{t=1}^T \mathcal{F}_t(y_t) \\
    \mathcal{F}_t(y_t) &= \KL[q(x_t ;\theta)||p(y_t, x_t | x_{t-1})] \numberthis
\end{align*}

This temporal factorization of the free energy means that the minimization at each timestep is independent of the others, and so we only need consider a single minimization of a single timestep to understand the solution, since all time-steps will be identical in terms of the solution method. Applying the linear gaussian assumptions of the Kalman filter, we can specify our generative model in terms of Gaussian distributions,
\begin{align*}
    p(y_t, x_t | x_{t-1}) &= p(y_t | x_t)p(x_t | x_{t-1}) \\
    &= \mathcal{N}(y_t; Cx_t, \Sigma_z)\mathcal{N}(x_t | Ax_{t-1}, \Sigma_x) \numberthis
\end{align*}

Since we know the posterior is Gaussian, it makes sense to also use a Gaussian distribution for the variational approximate distribution. Importantly, for predictive coding we make an additional assumption -- the Laplace Approximation -- which characterises the variance of this Gaussian as an analytic function of the mean, thus defining,
\begin{align*}
    q(x_t; \theta) = \mathcal{N}(x_t; \mu_t, \sigma(\mu)) \numberthis
\end{align*}
where $\theta = [\mu_t,\sigma(\mu_t)]$ are the parameters of the variational distribution -- in this case a mean and variance since we have assumed a Gaussian variational distribution. With the variational distribution and generative model precisely specified, it is now possible to explicitly evaluate the variational free energy for a specific time-step,
\begin{align*}
\mathcal{F}_t(y_t) &= \KL[q(x_t ;\theta)||p(y_t, x_t | x_{t-1})] \\
    &= -\E_{q(x_t;\theta}[\ln p(y_t, x_t | x_{t-1}] - \mathbb{H}[q(x_t;\theta)] \numberthis
\end{align*}
Where the second term is the entropy of the variational distribution. Since we are only interested in minimizing with respect to the mean $\mu_t$ and the expression for the entropy of a Gaussian does not depend on the mean, we can ignore this entropy term in subsequent steps. The key quantity is the `energy' term $\E_{q(x_t;\theta}[\ln p(y_t, x_t | x_{t-1}]$. Since the Laplace approximation ensures that most of the probability distribution is near the mode $\mu_t$ of the variational distribution, we can well approximate the expectation using a Taylor expansion to second order around the mode,
\begin{align*}
    \E_{q(x_t;\theta}[\ln p(y_t, x_t | x_{t-1}] &\approx \ln p(y_t, \mu_t | \mu_{t-1}) + \E[\frac{\partial p(y_t, x_t | x_{t-1})}{\partial x_t}|_{x_t = \mu_t}[x_t - \mu_t] + \E[\frac{\partial^2 p(y_t, x_t | x_{t-1})}{\partial x_t^2}|_{x_t = \mu_t}[x_t - \mu_t]^2 \\
    &= \ln p(y_t, \mu_t | \mu_{t-1}) + \frac{\partial p(y_t, x_t | x_{t-1})}{\partial x_t}|_{x_t = \mu_t}\underbrace{[\E[x_t] - \mu_t]}_{=0}+ \frac{\partial^2 p(y_t, x_t | x_{t-1})}{\partial x_t^2}|_{x_t = \mu_t}\underbrace{E[(x_t - \mu_t)^2]}_{=\sigma} \numberthis
\end{align*}

Since the first term vanishes as $\E[x_t] - \mu_t = \mu_t - \mu_t = 0$ and we can neglect the second term since it only depends on $\sigma$ and not $\mu$, then the only term that matters for the minimization is the first term $\ln p(y_t , \mu_t | \mu_{t-1})$. This means that we can write the overall optimization problem solved by predictive coding as,
\begin{align*}
    \underset{\mu_t}{arg min} \, \mathcal{F}_t{y_t} = \underset{\mu_t}{arg min} \, \ln p(y_t, \mu_t | \mu_{t-1}) \numberthis
\end{align*}
which is the same as the MAP optimization problem presented in equation 4. This means that ultimately the variational inference problem solved by predictive coding and the MAP estimation problem solved by the Kalman filter are the same although the interpretation of $\mu_t$ differs slightly -- from being a parameter of a Gaussian variational distribution versus simply a variable in the generative model -- the actual update rules involving $\mu_t$ are the same in both cases. Thus the differences between the MAP Kalman filtering objective and the variational predictive coding objective under the Laplace approximation are effectively only interpretational.

\end{document}